%% file: PaperForReview.tex
\documentclass[10pt,twocolumn,letterpaper]{article}
\usepackage[accsupp]{axessibility}

\usepackage[pagenumbers]{wacv} 

\usepackage{graphicx}
\usepackage{amsmath}
\usepackage{amssymb}
\usepackage{booktabs}
\usepackage{xcolor}
\usepackage{array}

\usepackage[pagebackref,breaklinks,colorlinks]{hyperref}

\usepackage[capitalize]{cleveref}
\crefname{section}{Sec.}{Secs.}
\Crefname{section}{Section}{Sections}
\Crefname{table}{Table}{Tables}
\crefname{table}{Tab.}{Tabs.}

\def\wacvPaperID{2116} 
\def\confName{WACV}
\def\confYear{2024}

\begin{document}

\title{Impact of Blur and Resolution on Demographic Disparities in 1-to-Many Facial Identification}

\author{Aman Bhatta$^1$  \quad\quad Gabriella Pangelinan$^2$ \quad\quad Michael C. King$^2$ \quad\quad Kevin W. Bowyer$^1$ \\\\
University of Notre Dame$^1$ \\Florida Institute of Technology$^2$ \\
{\tt\small \{abhatta,kwb\}@nd.edu, gpangelinan@my.fit.edu, michaelking@fit.edu }
}
\maketitle

\begin{abstract}

Most studies to date that have examined demographic variations in face recognition accuracy have analyzed 1-to-1 matching accuracy, using images that could be described as ``government ID quality”. 
This paper analyzes the accuracy of 1-to-many facial identification across demographic groups, and in the presence of blur and reduced resolution in the probe image as might occur in ``surveillance camera quality’’ images. 
Cumulative match characteristic curves (CMC) are not appropriate for comparing propensity for rank-one recognition errors across demographics, and so we use three metrics for our analysis: (1) the well-known d’ metric between mated and non-mated score distributions, and introduced in this work, (2) absolute score difference between thresholds in the high-similarity tail of the non-mated and the low-similarity tail of the mated distribution, and (3) distribution of (mated – non-mated rank-one scores) across the set of probe images. We find that demographic variation in 1-to-many accuracy does not entirely follow what has been observed in 1-to-1 matching accuracy.
Also, different from 1-to-1 accuracy, demographic comparison of 1-to-many accuracy can be affected by different numbers of identities and images across demographics.
More importantly, we show that increased blur in the probe image, or reduced resolution of the face in the probe image, can significantly increase the false positive identification rate.
And we show that the demographic variation in these high blur or low resolution conditions is much larger for male / female than for African-American / Caucasian.
The point that 1-to-many accuracy can potentially collapse in the context of processing ``surveillance camera quality’’ probe images against a ``government ID quality” gallery is an important one.

\end{abstract}

\input{fig_texs/intro_fig}

\section{Introduction}

Many studies to characterize demographic disparities in the accuracy of face recognition systems employed techniques designed to evaluate 1-to-1 match accuracy in the context of verification or authentication tasks, with only a few focusing on open-set 1-to-many identification scenarios. By all accounts, 1-to-many facial identification has been shown to be highly accurate in the case where the probe and gallery images are ``all government ID quality’' and the person in the probe image has one or more enrolled images \cite{Krishnapriya_TTS_2020, Grother_NIST_2018}.  %

In this work, we report experimental results investigating (a) demographic accuracy variation in 1-to-many face matching, and (b) impact of varying probe image quality on accuracy of 1-to-many face matching. 

In 1-to-many matching, an image of an unknown person (a ``probe'') is matched against enrolled images of many known persons (the ``gallery'') in order to find a candidate identity for the unknown person. 
In this paper, we focus on the use of 1-to-many matching to return the gallery image with the greatest similarity to the probe, the ``rank-one match.'' 
More generally, 1-to-many matching can also be used to return the top N matches, and / or to return matches above some threshold of similarity.
Such use-cases typically assume a human examiner is involved to decide among the candidate results, and experiments incorporating review of algorithm-selected candidate images by a human examiner are beyond the scope of this paper.

Contributions of this work include the following:
\begin{itemize}
\item Using ArcFace, MagFace and AdaFace, we compare 1-to-1 verification accuracy and 1-to-many identification accuracy across female / male and African-American / Caucasian. We find that the ranking of demographics for likelihood of rank-one false positive identification is not the same as the ranking for false match rate in 1-to-1 verification. This is due to 1-to-1 matching being based on a fixed threshold on similarity, whereas rank-one false positive identification is based on the difference between a probe’s rank-one mated and non-mated scores.

\item We introduce two metrics to supplement the well known d’ between mated and non-mated score distributions to estimate the propensity for 1-to-many matching to produce false positive identifications. The introduced metrics are: (1) absolute score difference between thresholds in the high-similarity tail of the non-mated and the low-similarity tail of the mated distribution, and (2) distribution of (mated – non-mated rank-one scores) across the set of probe images. These metrics allow an assessment of how close a distribution of scores is to generating false positive identifications even if no specific false positive identification is observed. The relative ranking of demographics is the same for the three metrics, but our analysis suggests that using d’ alone could miss subtle differences.

\item We create increasingly blurred versions of the original probe images in order to assess how blur in the probe image impacts false positive identification rate. We also create lower-resolution versions of the original probe images in order to assess how lower resolution in the probe image impacts false positive identification. For both increasing blur and lower resolution, there is a point at which one-to--many matching ``breaks” and the false positive identification rate increases dramatically. This suggests that 1-to-many search accuracy cannot be objectively evaluated without the context of (at least) the level of blur in the image and the resolution of the face.

\end{itemize}

\section{Related Work}

For a broad overview of issues related to demographic bias in biometrics, see the survey by Drozdowski et al. \cite{Drozdowski_TTS_2020}.
We briefly touch on related works in two areas: (1) demographic differences in face recognition accuracy and (2) 1-to-many matching for face recognition.

The earliest observation that accuracy varies across demographic groups may be the 2002 Face Recognition Vendor Test \cite{frvt}, 
which found lower accuracy for females than for males.
The well-known study by Klare et al. \cite{Klare2012} analyzed results from multiple algorithms and found that, ``The female, Black, and younger cohorts are more difficult to recognize for all matchers used in this study (commercial, nontrainable, and trainable)’’.
It is interesting that \cite{Klare2012} found that {\it training on demographic-balanced training data did not balance accuracy across demographics}.
These papers \cite{frvt,Klare2012} were before the rise of deep-CNN-based face matching. One recent study involving deep CNN matchers opposes the result that younger cohorts are more difficult to recognize \cite{Albiero_WACV_2020}. Another recent study agrees that the demographic balance of the training data does not lead to balanced accuracy across demographics \cite{Albiero_WACV_2020b}, meanwhile the study by Wu et al. \cite{wu_bmvc_2023} proposes a bias-aware cross-demographic validation methodology for bias factor investigation. Furthermore, a recent study demonstrates demographic disparities in accuracy at the facial detail attribute level too \cite{wu_cvpr_2023}.
The consensus across recent studies is that females, compared to males, have a worse impostor distribution and a worse genuine distribution and that African-American males have a worse impostor distribution compared to Caucasian males, but also a better genuine distribution \cite{Albiero_CVPRW_2020,Albiero_TIFS_2021,Krishnapriya_TTS_2020,Vangara_CVPRW_2019,Wu_2022,bhatta_wacvw_2023}.
All of the above studies are in the context of 1-to-1 matching.

\input{fig_texs/one2one_baseline}
Grother and colleagues have released NIST reports on 1-to-many facial identification \cite{Grother_NIST_2018} and on demographic effects \cite{Grother_NIST_2019,Grother2022}.
The report on 1-to-many identification \cite{Grother_NIST_2018} analyzes results from a number of matchers and large image sets.
They note the high 1-to-many rank-one accuracy of current matchers, the fact that accuracy decreases linearly with the number of enrolled identities, and that enrolling multiple images for an identity increases accuracy.
The report on demographic effects \cite{Grother_NIST_2019} notes results similar to those shown later in this paper for African-American / Caucasian, and for female / male.
They also discuss the specialized forms of 1-to-many matching that are designed to produce a list of candidates for review by a human analyst.
The most recent NIST report \cite{Grother2022} summarizes demographic comparisons across groups from various locations around the globe, and notes that demographic differences in false positive identification rate (FPIR) are larger than in false negative identification rate (FNIR), and that universal generalizations are difficult.
They also note that some commerical matchers do not implement 1-to-many matching in the straightforward manner \cite{Grother2022}, and so comparisons involving commercial matchers can be difficult. 
Krishnapriya et al \cite{Krishnapriya_TTS_2020} also look at 1-to-many matching accuracy, using the MORPH dataset, and also report that when the person in the probe image does have an enrolled image, the 1-to-many search is highly likely to return the correct identity.
Both \cite{Grother_NIST_2019} and \cite{Krishnapriya_TTS_2020} emphasize that a 1-to-many search when the person in the probe image does not have an enrolled image can by definition only return a false-positive identification (FPI).
Recent work by Drozdowski et al. \cite{Drozdowski_ICCVW_2021} examines the ``watchlist imbalance effect’’ that occurs in 1-to-many matching when different demographics represent different fractions of the watchlist.
They compare theoretical and empirical estimates of FPIR across demographics for balanced and unbalanced watchlists.
They also demonstrate that equitable 1-to-1 matching doesn’t necessarily ensure equitable 1-to-many matching. 

This paper goes beyond the works above in the following respects.
While the NIST reports use datasets of different quality images - mugshots, webcam and in-the-wild - they do not systematically examine quality dimensions associated with surveillance-camera images.
We perform systematic experiments examining the degree of blur and the degree of reduced resolution in the probe image, and how these factors impact 1-to-many accuracy.
Blur and resolution are two important elements in surveillance-camera quality images.
Our experiments use image datasets and matchers that in principle are available to the research community (some government datasets are not \cite{Grother_NIST_2019,Grother2022}), enabling reproducibility by other researchers.

\section{Dataset and Matcher}
MORPH \cite{morph_site, morph_paper} may be the best-known dataset in face aging research and is also widely used in studying demographic accuracy variation~\cite{Albiero_BMVC_2020, Albiero_CVPRW_2020, Albiero_WACV_2020,Drozdowski_ICCVW_2021, Georgopoulos_IJCV_2021, Krishnapriya_TTS_2020, Vangara_CVPRW_2019}.
Drozdowski et al.~\cite{Drozdowski_ICCVW_2021} note that MORPH is particularly appropriate for demographic  studies ``... due to its large size, relatively constrained image acquisition conditions, and the presence of ground-truth labels (from public records) for sex, race, and age of the subjects''.  
We use the same version of MORPH used in \cite{Albiero_TIFS_2021}, which has 35,276 images of 8,835 Caucasian males, 10,941 images of 2,798 Caucasian females, 56,245 images of 8,839 African-American males, and 24,855 images of 5,928 African-American females. 
The average number of images per identity is about 4 for Caucasian male and Caucasian female, 4.2 for African-American female and 6.4 for African-American male.

We use three open-source matchers: 1) combined margin model based on ArcFace loss~\cite{Deng_CVPR_2019} trained on Glint-360K(R100) \cite{An_ICCV_2021} with weights  at \cite{Insightface}, 2) MagFace \cite{Meng_cvpr_2021} trained on MS1MV2(R100) with weights  at \cite{MagFace}, and c) AdaFace \cite{Kim_CVPR_2022} trained on MS1MV2(R100) with weights at \cite{AdaFace}. The input to each network is an aligned face resized to 112x112, and the output is a 512-d feature vector that is matched using cosine similarity. Faces are detected and aligned using img2pose~\cite{Albiero_CVPR_2021}.
\input{table_texs/dprime_table}

\section{1-to-1 Matching Accuracy Across Demographics} \label{verification}

The top row of Figure~\ref{fig:one2one_baseline} shows the 1-to-1 impostor and genuine distributions. %
The impostor  contains similarity scores for all pairs of images representing different identities.
The genuine  contains similarity scores for all pairs of images representing the same identity. 
We also give the d' for separation of the impostor and genuine distributions \cite{d_prime_formula}:
\begin{equation}
 d' = \frac{|\mu_{G} - \mu_{I}|}{ \sqrt{\frac{\sigma^2_G + \sigma^2_I }{2}}}
\end{equation}
where $\mu_{G}$ and $\sigma_{G}$ are the mean and standard deviation for the genuine distribution, and $\mu_{I}$ and $\sigma_{I}$ for the impostor distribution.
Larger d' means greater separation between impostor and genuine distributions, implying greater matching accuracy.
(Using  d'  assumes that the distributions are reasonably approximated as Gaussian.)

Figure~\ref{fig:one2one_baseline} shows that  African-American male has the largest d', 
followed by Caucasian male, African-American female, and Caucasian female. 
However, the difference in d' for African-American female and Caucasian female is small.
This pattern of d' implies that if 1-to-1 matching was performed separately for each demographic, the lowest false non-match rate (FNMR) at a fixed false match rate (FMR) could be achieved for African-American male, followed by Caucasian male, African-American female, and Caucasian female. 
However, in the typical operational scenario, the same threshold on similarity score is used to classify all image pairs, regardless of demographics. 
Using a fixed threshold results in the different demographics having different FMRs, with Caucasian male having the lowest FMR, followed by African-American male, Caucasian female and African-American female.

\section{1-to-Many Accuracy Across Demographics in ``government ID" quality setting} \label{identification}

For our analyses of 1-to-many matching, each identity's most recent image is designated as its probe image and the older images as its enrolled images for the gallery. 
A few MORPH identities have just one image, and so have a probe but no enrolled images.
Given the high rank-one matching accuracy of modern matchers \cite{Grother_NIST_2018,Grother_NIST_2019,Krishnapriya_TTS_2020},
the size of MORPH, and the quality of images in MORPH, experiments will not generate large numbers of rank-one false positive identifications.  
However, we can make useful comparisons across demographics in at least three ways: (1) the d' separation of rank-one non-mated and mated score distributions, (2) separation in the FPIR-centric tails of the non-mated and mated distributions, and (3) the distribution of (mated – non-mated) score differences.  
Because of the straightforward analogy to 1-to-1 impostor and genuine distributions, we begin with rank-one non-mated and mated score distributions.

The 1-to-many mated distribution contains one score for each identity that has one or more enrolled images.
That score is the maximum similarity score of the probe to any enrolled image of the same identity.
The 1-to-many non-mated distribution also contains one score for each probe, and
that score is the maximum similarity of the probe to any enrolled image of a different identity.

The 1-to-many non-mated and mated distributions are shown in the bottom row of Figure~\ref{fig:one2one_baseline}. 
The 1-to-many non-mated and mated distributions are centered at higher similarity than the 1-to-1 impostor and genuine distributions. 
This is because the 1-to-many score for a probe is the maximum similarity score across a set of images, whereas in 1-to-1 impostor and genuine distributions include similarity scores for all image pairs.

\input{fig_texs/mated_nonmated}
The d'  for the 1-to-1 matching distributions and 1-to-many matching distributions in Figure~\ref{fig:one2one_baseline} are summarized in Table~\ref{tab:d-prime}, columns ``1-to-1'' and ``1-to-many original''.
As with the 1-to-1 distributions, African-American male has the largest d' for the 1-to-many distributions, followed by Caucasian male.
However, the order of the d' values flips for Caucasian female and African-American female, so that African-American female has the lowest d' in the 1-to-many matching results.
This is an example of how demographic differences in 1-to-many matching do not necessarily repeat differences observed in 1-to-1 matching.

Results in Table~\ref{tab:d-prime} suggest that, other factors equal, African-American males should have the lowest false positive identification rate (FPIR).
However, there are caveats.
One is that the number of enrolled identities and the number of images per identity vary across demographics. %
Another is that d' may not adequately reflect variations in skew in the tails of the  distributions. %
We address these two concerns in later sections.
Another concern is that operational scenarios are typically ``open-set'' in the sense that there is no knowledge of whether the identity in the probe image has any image in the gallery.
Every 1-to-many search using a probe that has no image in the gallery can only result in a FPI, as pointed out in \cite{Grother_NIST_2018,Krishnapriya_TTS_2020}.
For this reason, a general conclusion about the FPIR across demographics in an operational scenario requires knowledge of the rate of open-set searches across demographics.

\subsection{Comparison With Equal Identities and Images} 
The non-mated and mated distributions can be affected by the number of identities enrolled and the number of images per identity \cite{Grother_CVPR_2004,Grother_NIST_2018}.
Therefore, a comparison focused on difference due to demographic alone should have the same number of identities and images across the demographics.
In our dataset,  Caucasian female  has the least number of identities, 2,797.
So we randomly select 2,797 identities from each of the other demographics.
The most recent image of each identity is again used as the probe.
To equalize the number of enrolled images, the next most recent image is selected as the one enrolled image per identity.
This results in each of the demographics having 2,797 probe images and 2,797 gallery images.
Further, to check that the age difference between probe and enrolled image for an identity is approximately balanced across demographics, we checked the distribution of time between mated images.

The d' values for the non-mated and mated distributions based on the balanced image sets are given in Table~\ref{tab:d-prime} (``1-to-many balanced'' column).
The d' order across demographics is the same as for the original unbalanced datasets.
However, the differences between the demographics are smaller for the balanced dataset.

\subsection{Metric for Tails of Non-mated and Mated}\label{Metric_region}
As mentioned earlier, using d' assumes the distributions are reasonably approximated as Gaussian.
If the high-similarity tail of the non-mated and/or the low-similarity tail of the mated are skewed differently enough across demographics,  comparison based on d' could give a misleading impression of relative FPIR.
To address this, we create a metric that reflects the separation between the tails of the distributions without assuming the distributions are Gaussian.
We compute the distance between the similarity values representing the lower one-thousandth of the mated distribution and the upper one-thousandth of the non-mated distribution.
A higher value of this metric indicates greater separation in the relevant tails of the distributions. 

Values of this metric are shown in Table~\ref{tab:non-parametric}, computed both for the original dataset with varying numbers of identities and images across demographics, and for the balanced dataset.
Note that the order of the demographics is the same for  unbalanced and  balanced datasets.
However, the difference between African-American male and Caucasian male is  minimal for the balanced dataset.
This again illustrates the need to make a more detailed analysis than d' alone.

\input{table_texs/non_parametric}

\input{fig_texs/blur_demo}

\input{fig_texs/gaussian}

\subsection{Distribution of (Mated - Non-mated) Difference}
For a given probe image, a negative (mated - non-mated) score difference represents an FPI,
and larger positive differences represent  further distance from possible FPI.
In this sense, the distribution of (mated - non-mated) score differences is a more direct indication of potential for FPI.
The distributions of (mated - non-mated) differences for the balanced datasets  
are shown in Figure~\ref{fig:mated_nonmated}.
Distributions for the  male demographics are relatively similar. 
Distributions for the  female demographics are also relatively similar.
The larger demographic difference is between male / female, and the difference between African-American / Caucasian of the same gender is smaller.
\input{fig_texs/downsample_demo}
\input{fig_texs/downsample}

\section{1-to-Many Accuracy Across Demographics in ``surveillance" setting} 

Results in earlier sections are based on both the probe image and the enrolled images both being ``government ID quality''.
However, one important scenario for the use of 1-to-many facial identification involves a ``surveillance camera'' probe image matched against government ID-quality enrolled images. 
Probe images derived or extracted from surveillance media may be lower quality, with respect to image blur, resolution of the face, and other factors.
In this section we explore how \textit{blur in the probe} image impacts 1-to-many accuracy, and subsequently, we explore the impact of the \textit{probe having lower resolution}.

\subsection{Impact of Blur in Probe Image}\label{blur_exp}
\vspace{-0.5em}
For this experiment, we used Gaussian smoothing of $\sigma = 1, 2, 3, 4, 5$  to create a sequence of increasingly blurred versions of the probe images (the same Gaussian blur settings as in \cite{Robbins_CVPRW_2022}). Sample probe images are shown in Figure \ref{fig:blur_demo}.
We repeated the 1-to-many matching experiment done with the original probe images with each of the increasing blurred versions of the probe images.
The resulting 1-to-many mated and nonmated distributions are shown in Figure \ref{fig:gaussian_blur}.
Increasing blur in the probe image has no large effect on the nonmated distribution, but causes the mated distribution to move to lower similarity.
As an example, Figure \ref{gb} shows that, with ArcFace as the chosen face matcher, at a blur level of $\sigma=5$, both the Caucasian female and African-American female demographics display false positive identification rates of over 1\% for probe images. These findings imply that estimating the level of blur in probe images may be a crucial quality control measure for any 1-to-many search result.
\input{fig_texs/fpir_plots}

The results consistently align across all three matchers utilized in this study — ArcFace, AdaFace, and MagFace. This finding is particularly intriguing given that the AdaFace and MagFace loss functions are intentionally designed to consider image quality during training, showing promise in 1-to-1 matching scenarios. However, it's worth noting that these promising results in 1-to-1 matching do not seamlessly extend to 1-to-many, as illustrated in Figure \ref{fig:gaussian_blur}.
\vspace{-0.5em}
\subsection{Impact of Lower-Resolution  Probe Image}
Faces cropped from frames of surveillance video may have lower spatial resolution than  government ID face images.
To explore the impact of having a lower-resolution probe face image, we used bicubic interpolation to create $84 \times 84$, $56 \times 56$, $42 \times 42$ and $28 \times 28$ versions of the original $224 \times 224$ probe images. Sample images of this sequence are shown in Figure \ref{fig:downsample_demo}.

The different resolution versions of the probe images are all converted to $112 \times 112$ for input to the CNN.
The 1-to-many matching experiment is repeated for each resolution of the probe images, and the results are shown in Figure \ref{fig:downsample}.
The nonmated distribution is relatively unaffected by the resolution of the probe image before it is input to the network.
But lower resolution in the probe image causes the mated distribution to shift to lower similarity.
As an example, Figure \ref{lr} shows that, with ArcFace as the chosen face matcher, for $28 \times 28$ versions of probe images, the female demographics exhibit a false positive identification rate of more than 10\%. 
It is important to acknowledge that the experimental setup used in this study provided the face embedding network with more prior information than what is commonly encountered in real-world scenarios. Specifically, the images utilized were pre-aligned and labeled, with the underlying assumption that the ``face is certainly in the image." In practice, face detection algorithms may discard images that are too small or cannot be aligned, unless a human operator annotates the face image. Furthermore, in alignment with our observations from the blur experiment, as presented in Section \ref{blur_exp}, we find a consistent pattern in the results across all three matchers for this experiment as well. 

\section{Conclusions and Discussion}

We focus on accuracy of 1-to-many matching in terms of the false positive identification rate for the rank-one match of a probe image against a set of gallery images that contains one or more other images of the same identity as the probe (``closed-set'' matching).
We do not consider schemes in which the 1-to-many search returns all matches above a similarity threshold or the top N matches.
For a series of results based on both the probe image and the enrolled images being government ID-quality, we use three different approaches to assess the propensity for false positive identification in 1-to-many search.
We show that the relative ranking across demographics for 1-to-many matching is different than the ranking for 1-to-1 matching.
This is because 1-to-1 matching uses a fixed similarity threshold across demographics, while rank-one 1-to-many matching results in a false positive identification when the rank-one nonmated match is higher similarity than the rank-one mated match.
We also point out that comparison of 1-to-many accuracy across demographics should be done with equal number of identities and images/identity across demographics.
And we show that using only the d' for separation of mated and non-mated distributions may not be sufficient to appropriately compare accuracy across demographics.

We show that increasing blur in the probe image causes the mated distribution to shift to lower similarity and thereby increases the FPIR.
Similarly, we  show that lower spatial resolution in the probe image causes the mated distribution to shift to lower similarity and thereby increase the FPIR.
The high accuracy of modern CNN-based matchers, when both probe and enrolled images are of government ID quality \cite{Krishnapriya_TTS_2020,Grother_NIST_2019}, can degrade dramatically if the probe image exhibits substantial blur or low resolution on the face, as is often the case in real-world ``surveillance" settings.

Image blur level and spatial resolution of the cropped face are quality factors that are readily estimated.
Our results suggest that these factors should be considered when assessing the usefulness of a probe for 1-to-many search.
Our analysis here, employing three matchers (ArcFace, AdaFace, and MagFace) and utilizing MORPH as a test dataset, is an example of the type of analysis that could be done for a particular operational scenario's matcher and face dataset in order to establish quality thresholds for probe images used in facial identification.

{\small
\bibliographystyle{ieee_fullname}
\bibliography{egbib}
}

\end{document}

%% file: fig_texs/intro_fig.tex
\begin{figure}[tb]
    \raggedright
    \includegraphics[width=.95\linewidth]{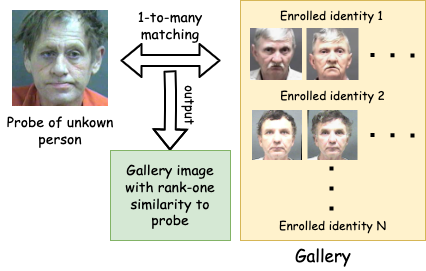}
    \caption{Does 1-to-many rank-one match error rate vary across demographics? 
In 1-to-many identification, an image of a person with unknown identity (the probe) is matched against a list of persons with known identity (the gallery) to find a candidate identity.}
\vspace{-1.0em}
\label{fig:intro}
\end{figure}

%% file: fig_texs/one2one_baseline.tex
\begin{figure*}[ht!]
  \begin{subfigure}[b]{1\linewidth}
    \centering
      \begin{subfigure}[b]{0.24\linewidth}
        \centering
          \includegraphics[width=1\linewidth]{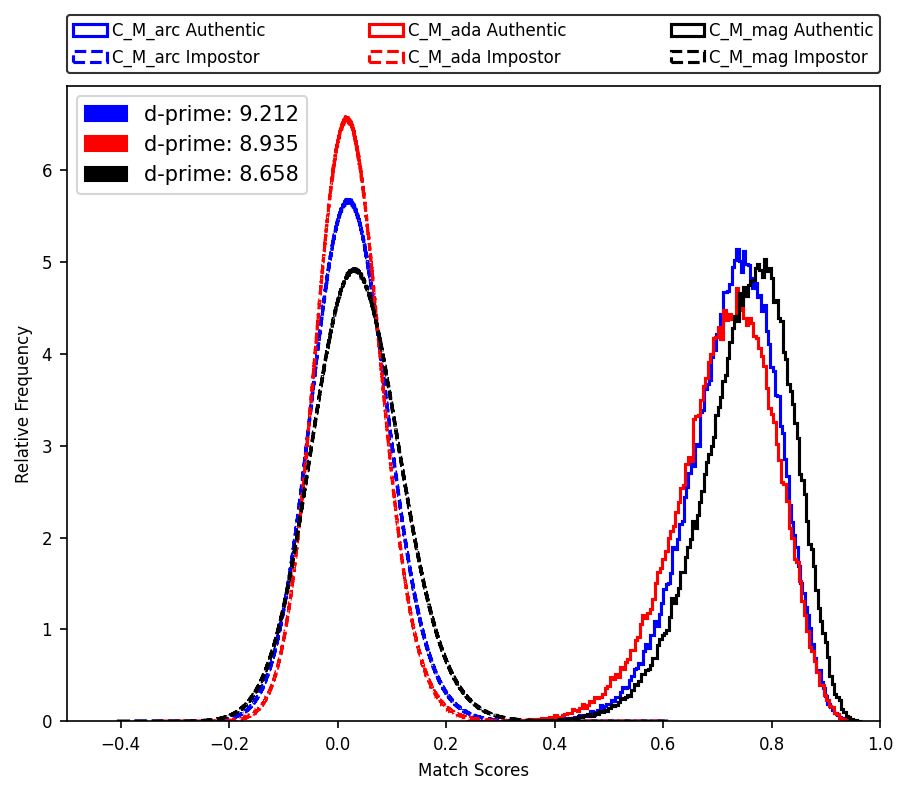}
      \end{subfigure}
      \hfill
      \begin{subfigure}[b]{0.24\linewidth}
        \centering
          \includegraphics[width=1\linewidth]{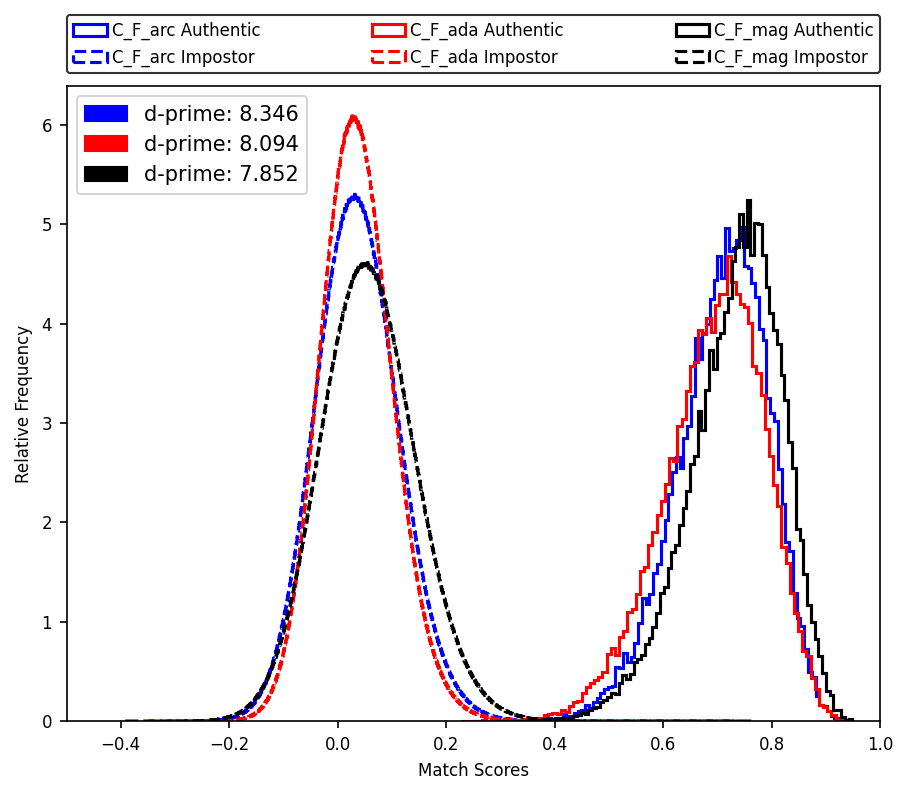}
      \end{subfigure}
      \hfill
      \begin{subfigure}[b]{0.24\linewidth}
        \centering
          \includegraphics[width=1\linewidth]{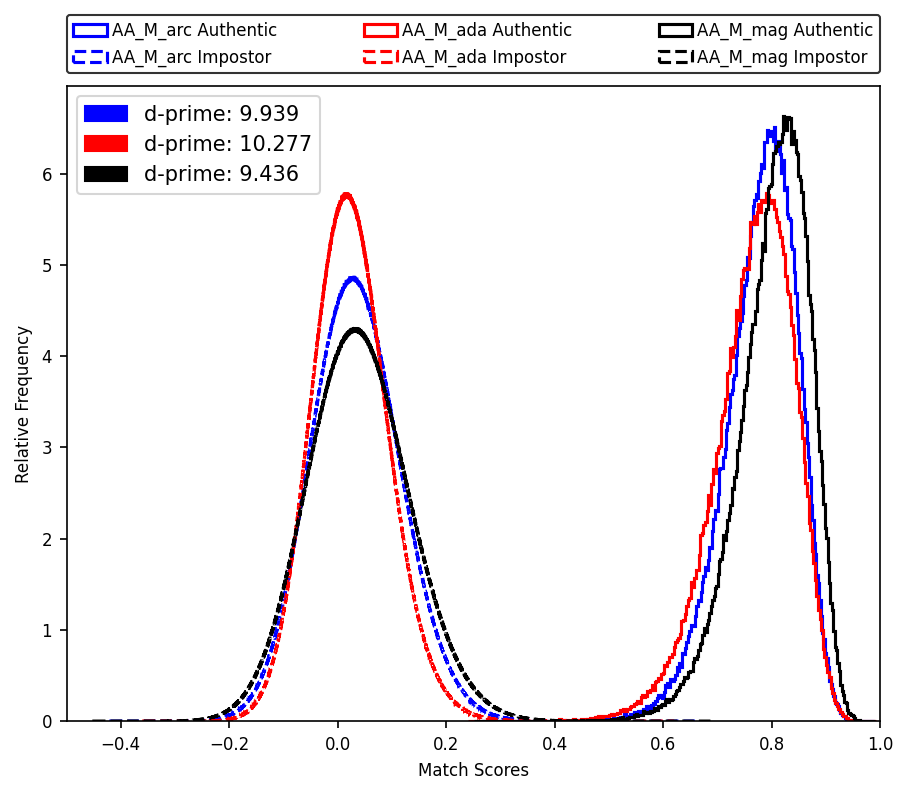}
      \end{subfigure}
      \hfill
      \begin{subfigure}[b]{0.24\linewidth}
        \centering
          \includegraphics[width=1\linewidth]{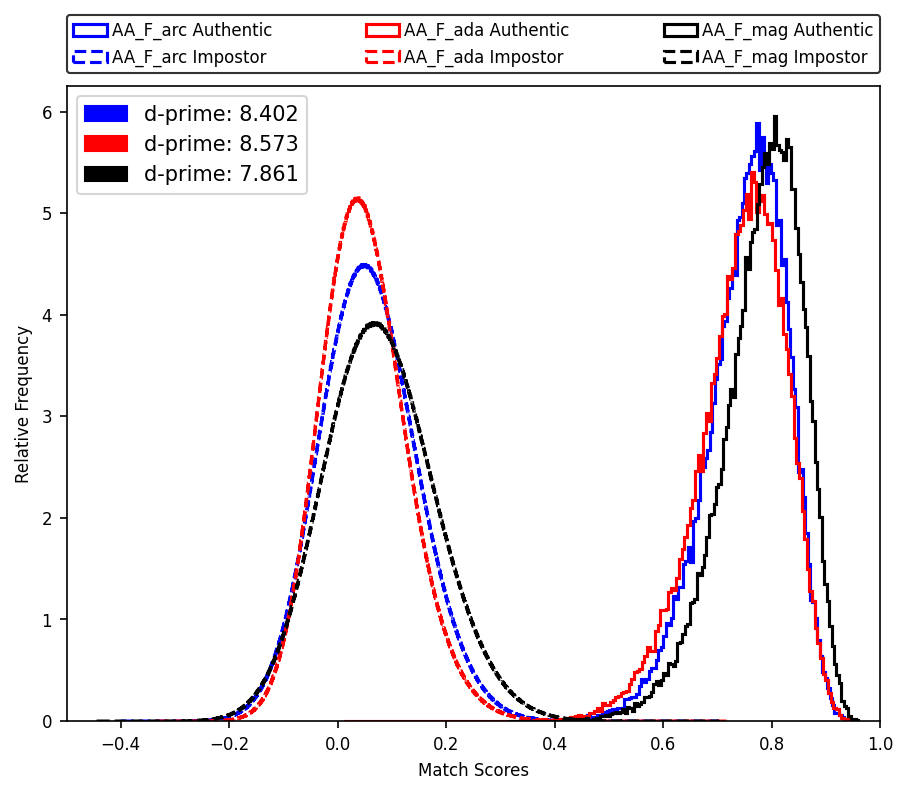}
      \end{subfigure}
  \end{subfigure}
  \hfill
  \begin{subfigure}[b]{1\linewidth}
  \centering
      \begin{subfigure}[b]{0.24\linewidth}
        \centering
          \includegraphics[width=1\linewidth]{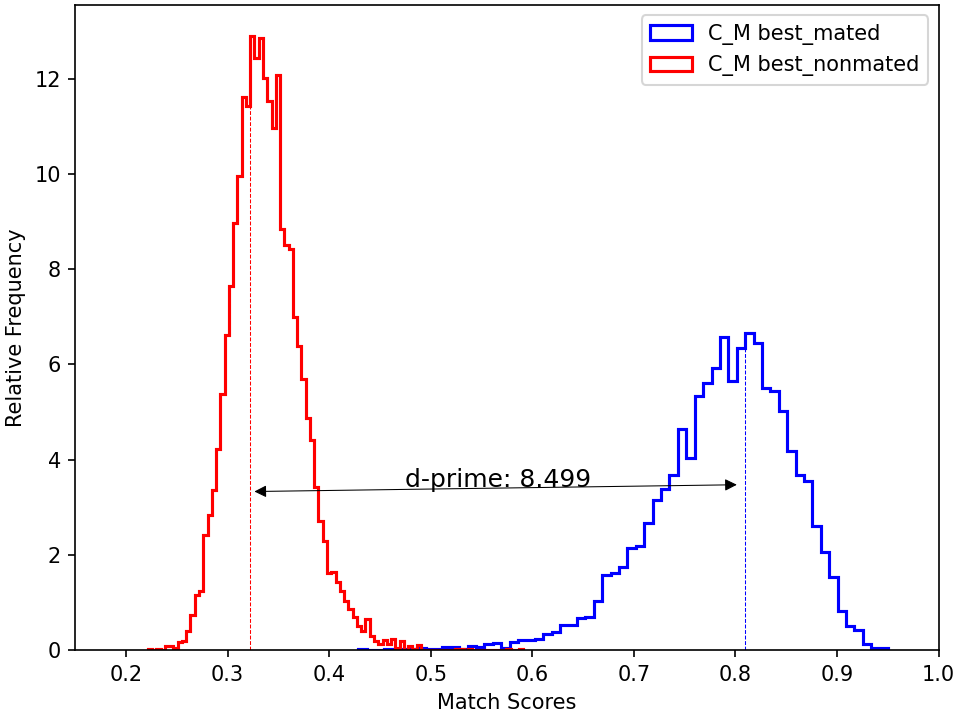}
            \caption{Caucasian Male}
            \vspace{-0.5em}
      \end{subfigure}
      \hfill
      \begin{subfigure}[b]{0.24\linewidth}
        \centering
          \includegraphics[width=1\linewidth]{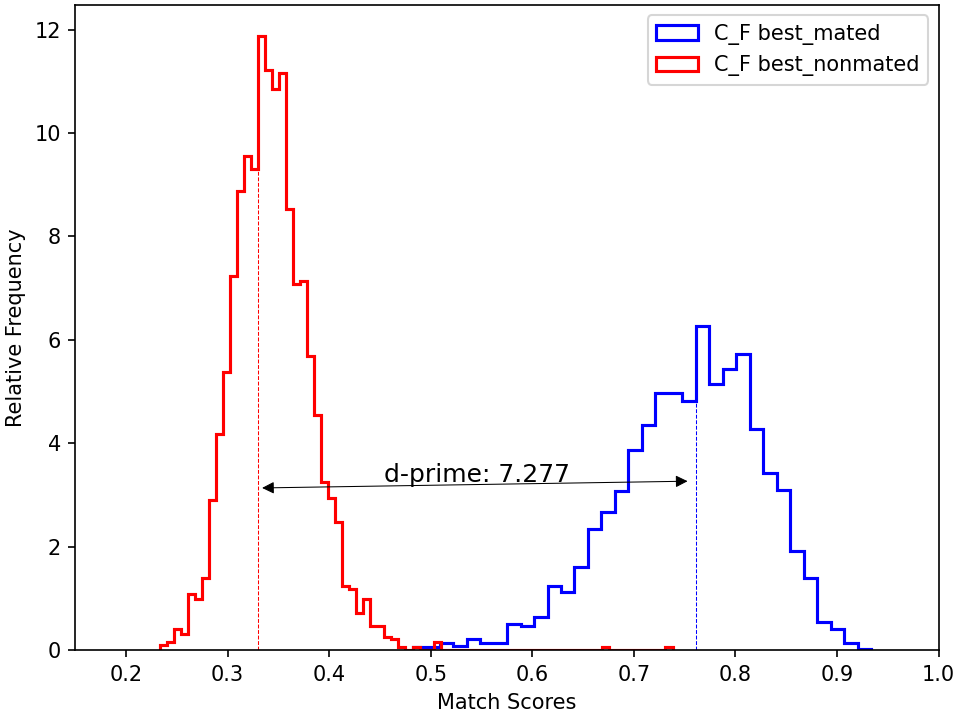}
         \caption{Caucasian Female}
         \vspace{-0.5em}
      \end{subfigure}
      \hfill
      \begin{subfigure}[b]{0.24\linewidth}
        \centering
          \includegraphics[width=1\linewidth]{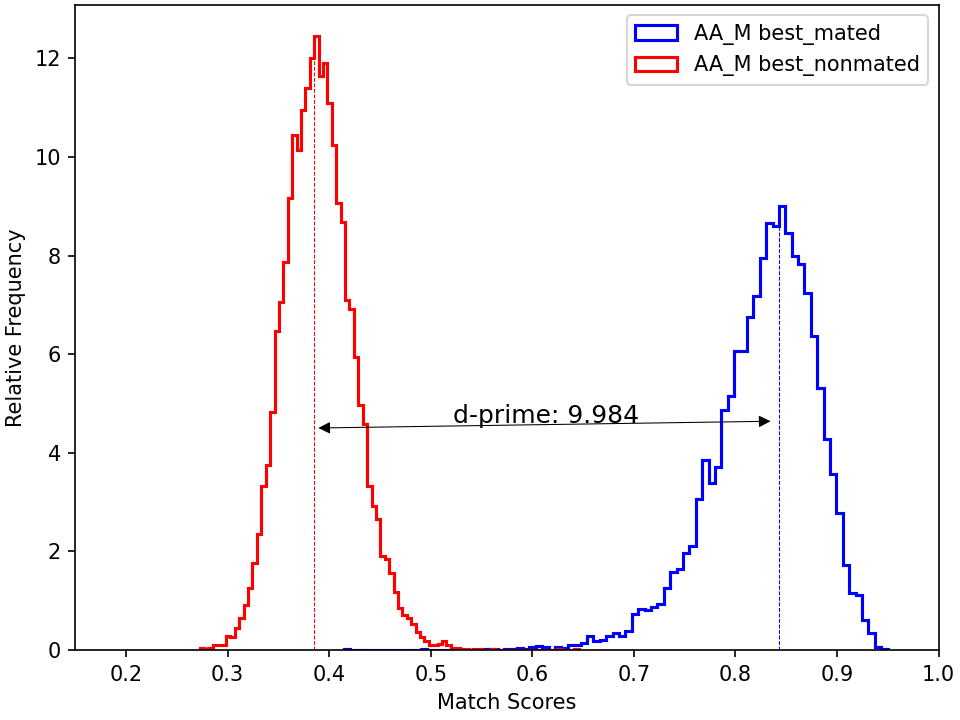}
            \caption{African-American Male}
            \vspace{-0.5em}
      \end{subfigure}
      \hfill
      \begin{subfigure}[b]{0.24\linewidth}
        \centering
          \includegraphics[width=1\linewidth]{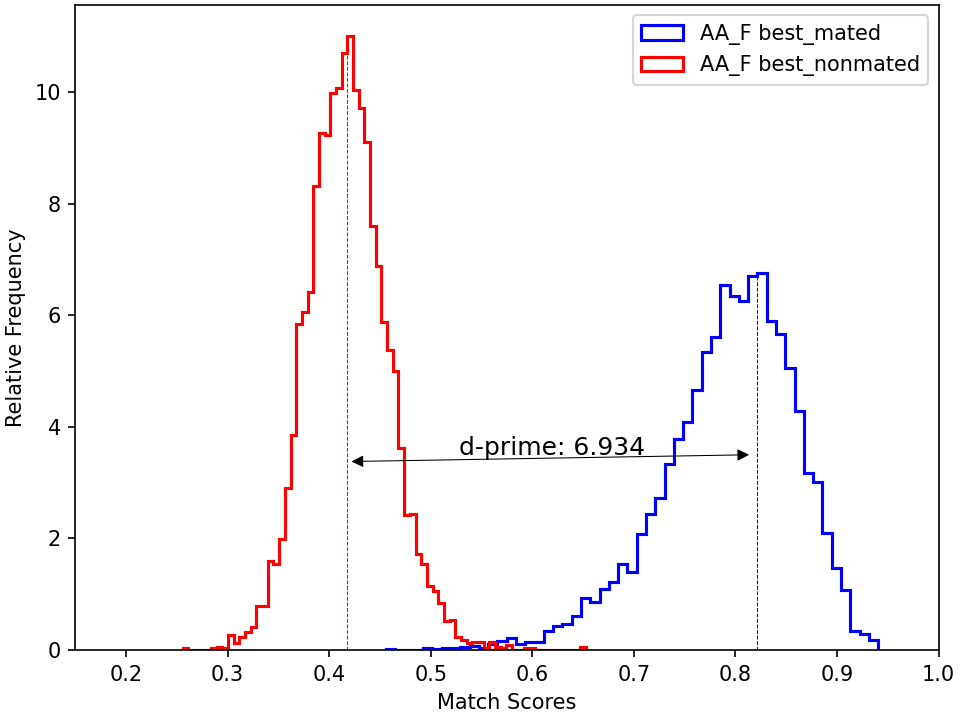}
            \caption{African-American Female}
            \vspace{-0.5em}
      \end{subfigure}
  \end{subfigure}
  \caption{Baseline 1-to-1 (verification) and 1-to-many (identification) distributions. Top row shows 1-to-1 impostor and genuine distributions; d-prime is greatest for African-American Male and lowest for Caucasian Female. Bottom row shows 1-to-many mated and non-mated distributions for ArcFace (results for AdaFace and MagFace in Table \ref{tab:d-prime}) ; d-prime is greatest (lowest false positive identification rate) for African-American Male and lowest (highest false positive identification rate) for African-American Female. In the legend of figures in the top row, the labels ``arc," ``ada," and ``mag" correspond to the results obtained using the ArcFace, AdaFace, and MagFace loss functions, respectively.}
  \label{fig:one2one_baseline}
\end{figure*}

%% file: table_texs/dprime_table.tex
\setlength\extrarowheight{2pt}
\begin{table*}[h!]
\centering
\resizebox{0.66\linewidth}{!}{%
\begin{tabular}{|l|ccc|ccc|ccc|}
\cline{1-10}
\multicolumn{1}{|c|}{\textbf{}} & \multicolumn{3}{c|}{\textbf{ArcFace}} & \multicolumn{3}{c|}{\textbf{AdaFace}} & \multicolumn{3}{c|}{\textbf{MagFace}} \\
\cline{2-10}
\multicolumn{1}{|c|}{\textbf{Group}} &
  \textbf{1-to-1} &
  \textbf{\begin{tabular}[c]{@{}c@{}}1-to-many\\ original\end{tabular}} &
  \textbf{\begin{tabular}[c]{@{}c@{}}1-to-many\\ balanced\end{tabular}} &
  \textbf{1-to-1} &
  \textbf{\begin{tabular}[c]{@{}c@{}}1-to-many\\ original\end{tabular}} &
  \textbf{\begin{tabular}[c]{@{}c@{}}1-to-many\\ balanced\end{tabular}} &
  \textbf{1-to-1} &
  \textbf{\begin{tabular}[c]{@{}c@{}}1-to-many\\ original\end{tabular}} &
  \textbf{\begin{tabular}[c]{@{}c@{}}1-to-many\\ balanced\end{tabular}} \\
  \cline{1-10}
\textbf{AA M}                 & 9.94       & 9.98       & 9.55       &  10.27          &    9.54        & 8.98           &  9.44          &  9.23          & 9.10           \\
\textbf{C M}                  & 9.21       & 8.50       & 8.43       & 8.93           &   8.12         & 8.07           &   8.66         & 7.99            & 7.91           \\
\textbf{C F}                  & 8.35       & 7.28       & 6.84       &  8.09          &   6.84         &  \textcolor[HTML]{FD6864} {6.40}        &   7.85         &  6.54          &  6.17          \\
\textbf{AA F}                 & 8.40       & 6.93       & 6.82       &  8.57          &    6.44        &  \textcolor[HTML]{FD6864} {6.41}       &   7.86         &  6.07          &  5.95         \\
\cline{1-10}
\end{tabular}
}
\vspace{0.25cm}
\caption{d' Comparisons Across Demographics.
d' for 1-to-1 impostor and genuine, 1-to-many non-mated and mated with original unbalanced dataset, and 1-to-many non-mated and mated with demographics balanced on number of identities and images.}
\label{tab:d-prime}
\end{table*}

%% file: fig_texs/mated_nonmated.tex
\begin{figure*}[ht!]
  \begin{subfigure}[b]{1\linewidth}
    \centering
      \begin{subfigure}[b]{0.33\linewidth}
        \centering
          \includegraphics[width=1\linewidth]{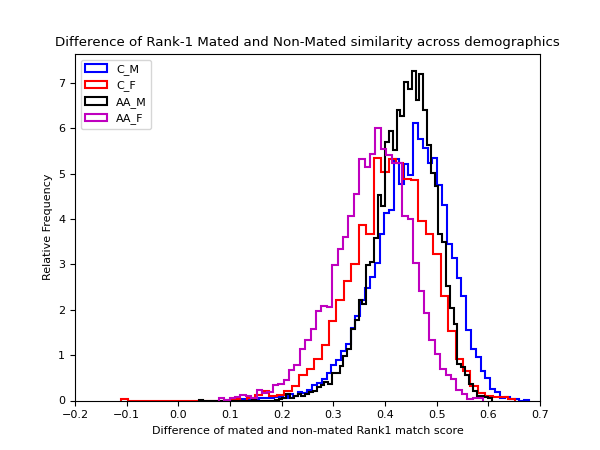}
          \caption{ArcFace}
      \end{subfigure}
      \hfill
      \begin{subfigure}[b]{0.33\linewidth}
        \centering
          \includegraphics[width=1\linewidth]{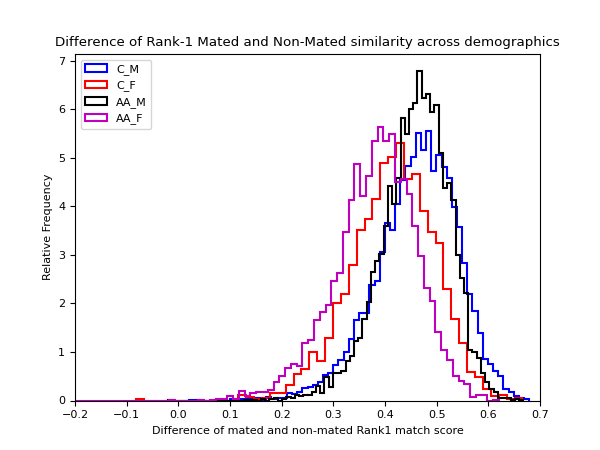}
          \caption{AdaFace}
      \end{subfigure}
      \hfill
      \begin{subfigure}[b]{0.33\linewidth}
        \centering
          \includegraphics[width=1\linewidth]{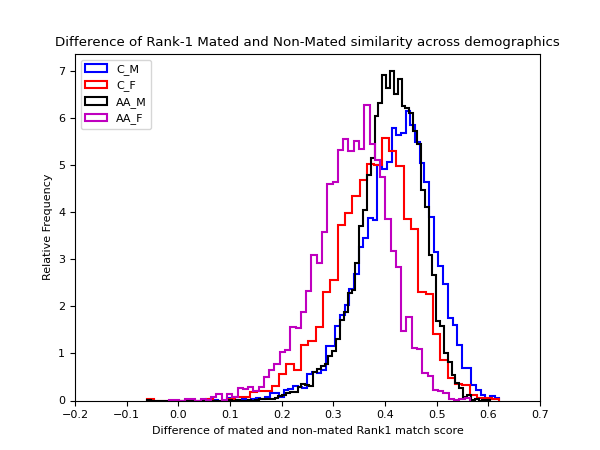}
          \caption{MagFace}
      \end{subfigure}
  \end{subfigure}
  \caption{(mated - non-mated) distributions by demographic for three face matchers. The X-axis represents the ``Relative Frequency",  while the Y-axis represents the ``Difference of mated and non-mated Rank1 match score".}
  \label{fig:mated_nonmated}
\end{figure*}

%% file: table_texs/non_parametric.tex
\begin{table}[t]
\centering
\resizebox{1\columnwidth}{!}{%
\begin{tabular}{|l|cc|cc|cc|}
\cline{1-7}
& \multicolumn{2}{c|}{\textbf{ArcFace}} & \multicolumn{2}{c|}{\textbf{AdaFace}} & \multicolumn{2}{c|}{\textbf{MagFace}} \\
\cline{2-7}
\textbf{Group} &
  \begin{tabular}[c]{@{}c@{}}$\Delta$\\ original\end{tabular} &
  \begin{tabular}[c]{@{}c@{}}$\Delta$\\ balanced\end{tabular} &
  \begin{tabular}[c]{@{}c@{}}$\Delta$\\ original\end{tabular} &
  \begin{tabular}[c]{@{}c@{}}$\Delta$\\ balanced\end{tabular} &
  \begin{tabular}[c]{@{}c@{}}$\Delta$\\ original\end{tabular} &
  \begin{tabular}[c]{@{}c@{}}$\Delta$\\ balanced\end{tabular} \\
  \cline{1-7}
\textbf{AA M} & 0.073             & 0.060            &   0.056                &  0.048                &  0.034                 & 0.029                 \\
\textbf{C M}  & 0.010             & 0.059            &  -0.027                 &  0.021                &     -0.023              & -0.048                 \\
\textbf{C F}  & -0.005            & 0.018            &   -0.038                & \textcolor[HTML]{FD6864} {-0.068}             &    -0.060             & \color[HTML]{FD6864}-0.123                 \\
\textbf{AA F} & -0.059            & -0.044           &    -0.096               &  \textcolor[HTML]{FD6864} {-0.058}               &  -0.101                 & \color[HTML]{FD6864}-0.083                \\
\cline{1-7}
\end{tabular}%
}
\vspace{0.25cm}
\caption{Non-parametric ``Recognition Power'' Across Demographics.
 $\Delta$ is the distance between the 1-in-1000 threshold in the high-similarity tail of the non-mated distribution and the low-similarity tail of the mated. Balanced indicates that the number of identities and number of enrolled images per identity are the same across demographic groups.}
\vspace{-0.5em}
\label{tab:non-parametric}
\end{table}

%% file: fig_texs/blur_demo.tex
\begin{figure*}[!ht]
  \begin{subfigure}[b]{1\linewidth}
    \centering
      \begin{subfigure}[b]{0.16\linewidth}
        \centering
          \includegraphics[width=1\linewidth]{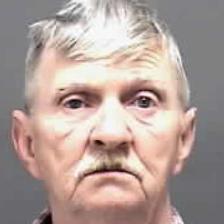}
          \caption{$\sigma =$ none }
      \end{subfigure}
      \hfill
      \begin{subfigure}[b]{0.16\linewidth}
        \centering
          \includegraphics[width=1\linewidth]{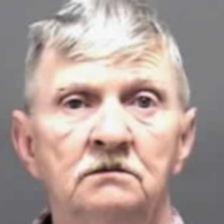}
          \caption{$\sigma = 1$ }
      \end{subfigure}
      \hfill
      \begin{subfigure}[b]{0.16\linewidth}
        \centering
          \includegraphics[width=1\linewidth]{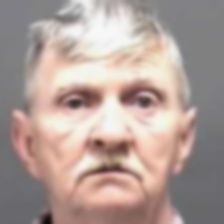}
          \caption{$\sigma = 2$ }
      \end{subfigure}
      \hfill
      \begin{subfigure}[b]{0.16\linewidth}
        \centering
          \includegraphics[width=1\linewidth]{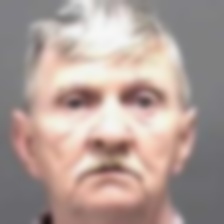}
          \caption{$\sigma = 3$ }
      \end{subfigure}
      \hfill
      \begin{subfigure}[b]{0.16\linewidth}
        \centering
          \includegraphics[width=1\linewidth]{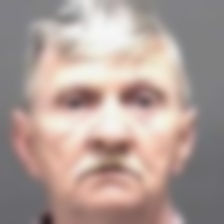}
          \caption{$\sigma = 4$ }
      \end{subfigure}
      \hfill
      \begin{subfigure}[b]{0.16\linewidth}
        \centering
          \includegraphics[width=1\linewidth]{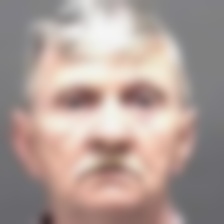}
          \caption{$\sigma = 5$ }
      \end{subfigure}
  \end{subfigure}
  \caption{Sample probe images with increasing Gaussian blur levels.}
  \label{fig:blur_demo}
\end{figure*}

%% file: fig_texs/gaussian.tex
\begin{figure*}[ht!]
\centering
  \begin{subfigure}[b]{1\linewidth}
    \centering
      \begin{subfigure}[b]{0.245\linewidth}
        \centering
          \includegraphics[width=1\linewidth]{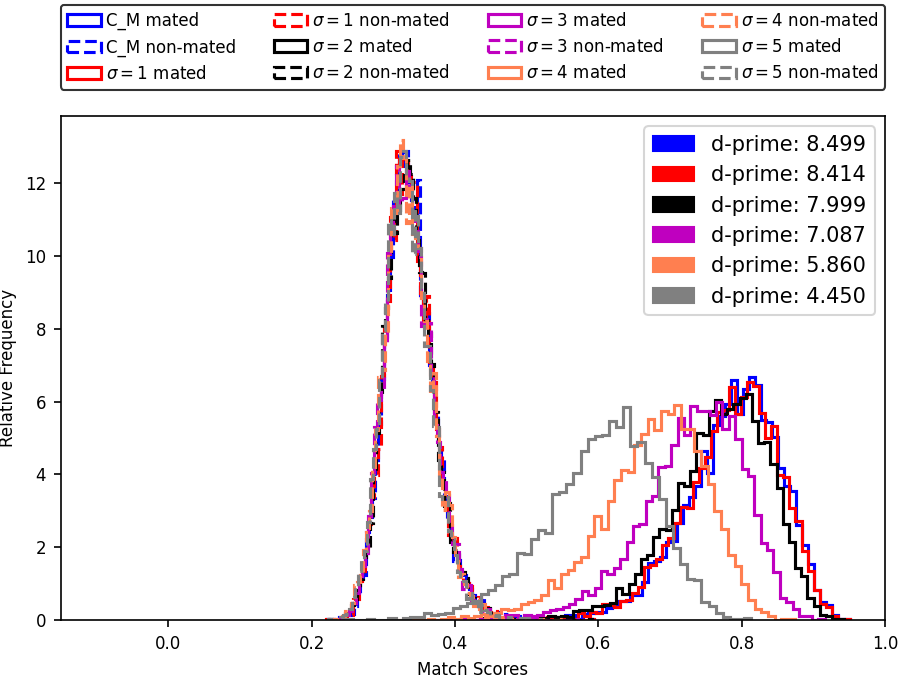}
      \end{subfigure}
      \hfill
      \begin{subfigure}[b]{0.245\linewidth}
        \centering
          \includegraphics[width=1\linewidth]{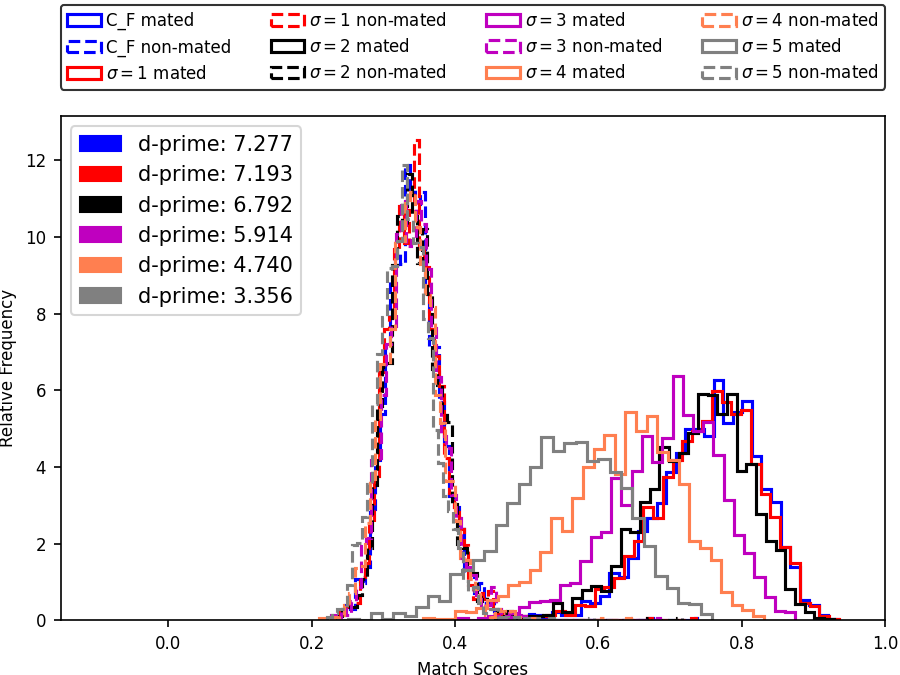}
      \end{subfigure}
      \hfill
      \begin{subfigure}[b]{0.245\linewidth}
        \centering
          \includegraphics[width=1\linewidth]{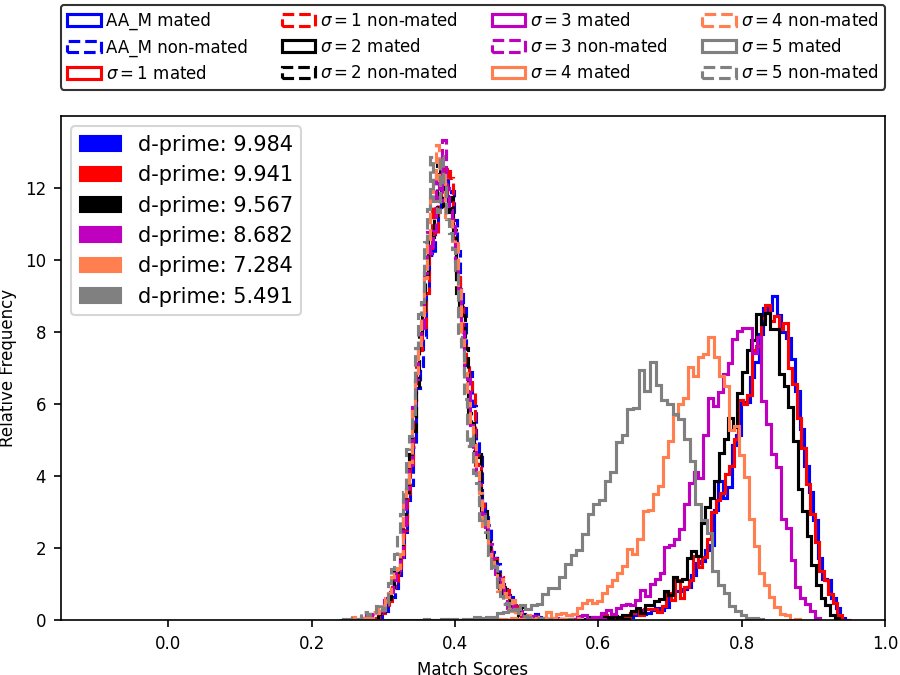}
      \end{subfigure}
      \hfill
      \begin{subfigure}[b]{0.245\linewidth}
        \centering
          \includegraphics[width=1\linewidth]{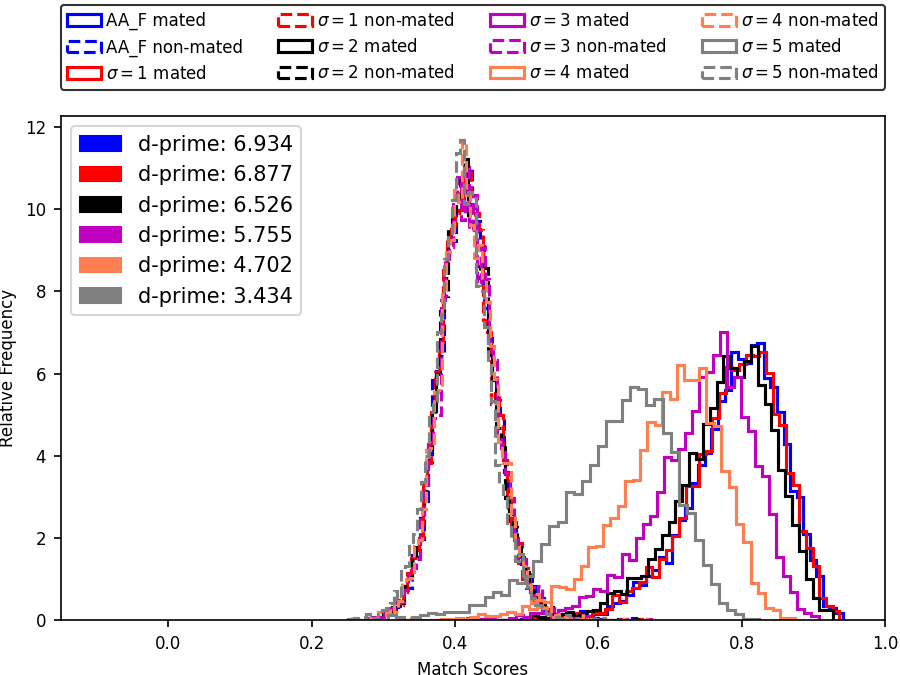}
      \end{subfigure}
  \end{subfigure}

  \vspace{10pt}
  
  \begin{subfigure}[b]{1\linewidth}
    \centering
      \begin{subfigure}[b]{0.245\linewidth}
        \centering
          \includegraphics[width=1\linewidth]{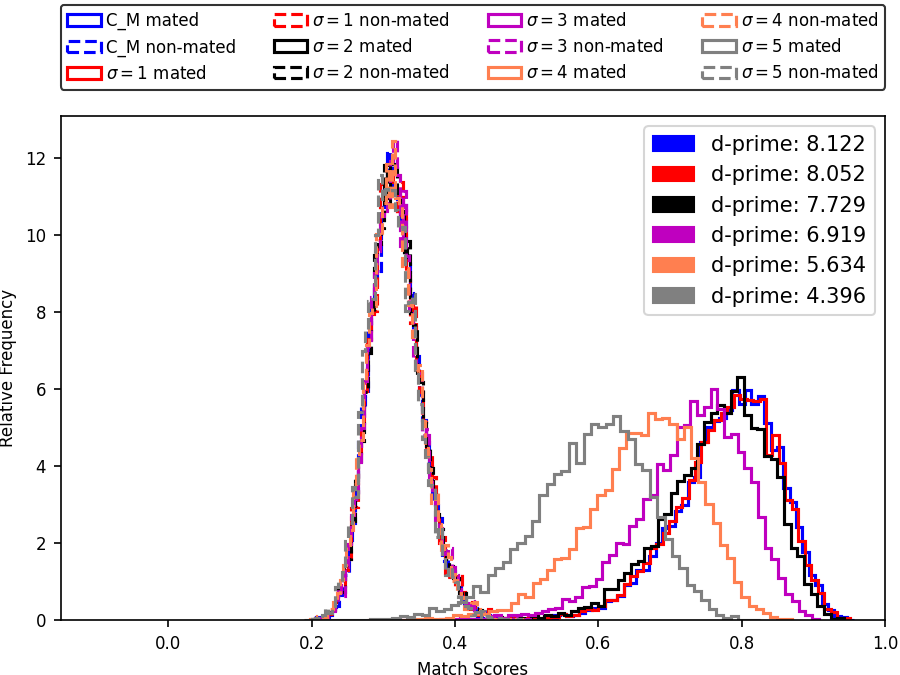}
      \end{subfigure}
      \hfill
      \begin{subfigure}[b]{0.245\linewidth}
        \centering
          \includegraphics[width=1\linewidth]{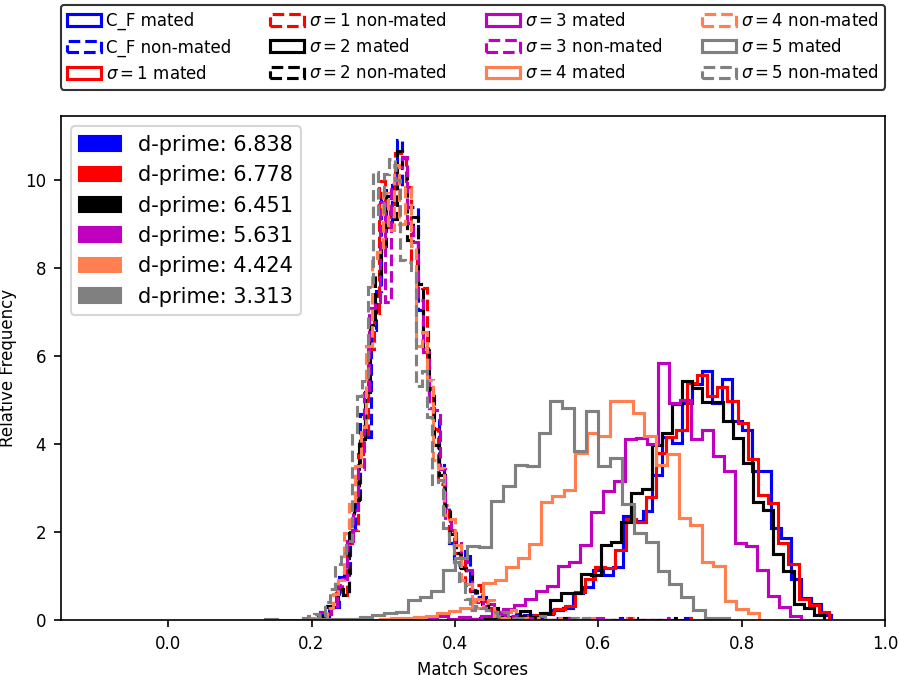}
      \end{subfigure}
      \hfill
      \begin{subfigure}[b]{0.245\linewidth}
        \centering
          \includegraphics[width=1\linewidth]{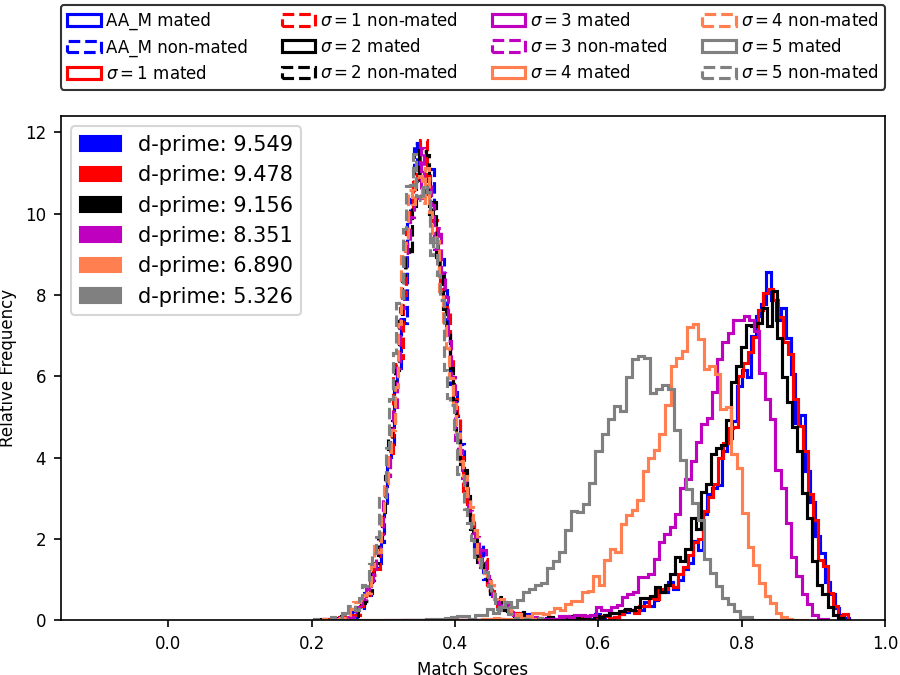}
      \end{subfigure}
      \hfill
      \begin{subfigure}[b]{0.245\linewidth}
        \centering
          \includegraphics[width=1\linewidth]{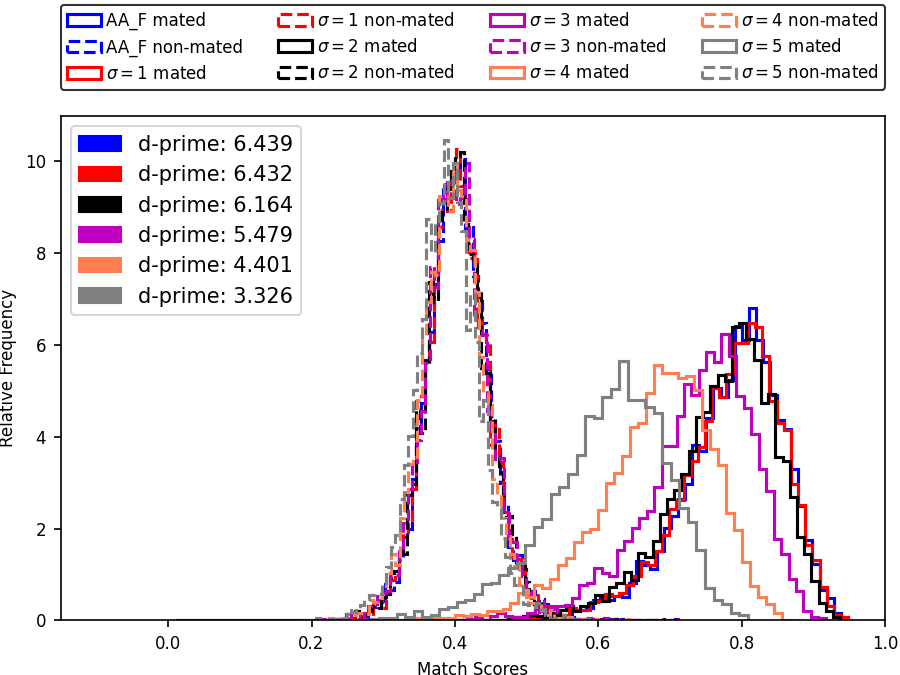}
      \end{subfigure}
  \end{subfigure}

  \vspace{10pt}
  
  \begin{subfigure}[b]{1\linewidth}
    \centering
      \begin{subfigure}[b]{0.245\linewidth}
        \centering
          \includegraphics[width=1\linewidth]{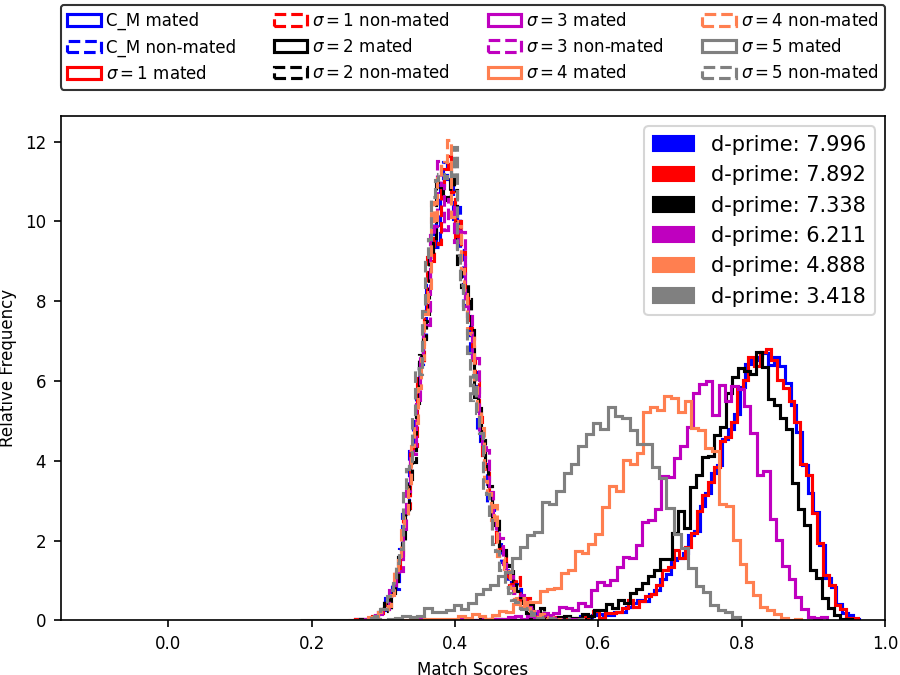}
          \caption{Caucasian Male}
      \end{subfigure}
      \hfill
      \begin{subfigure}[b]{0.245\linewidth}
        \centering
          \includegraphics[width=1\linewidth]{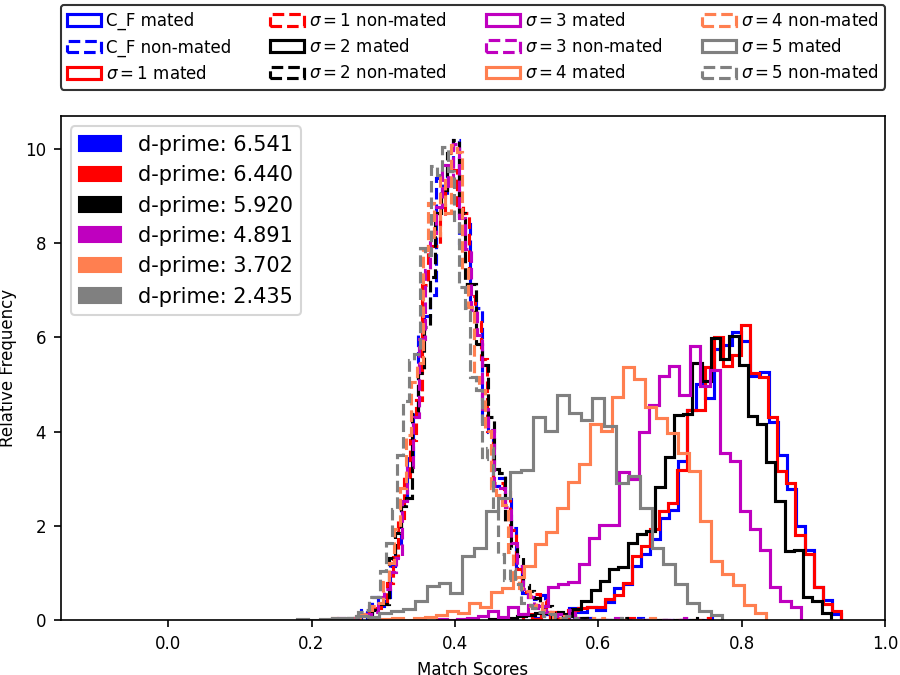}
          \caption{Caucasian Female}
      \end{subfigure}
      \hfill
      \begin{subfigure}[b]{0.245\linewidth}
        \centering
          \includegraphics[width=1\linewidth]{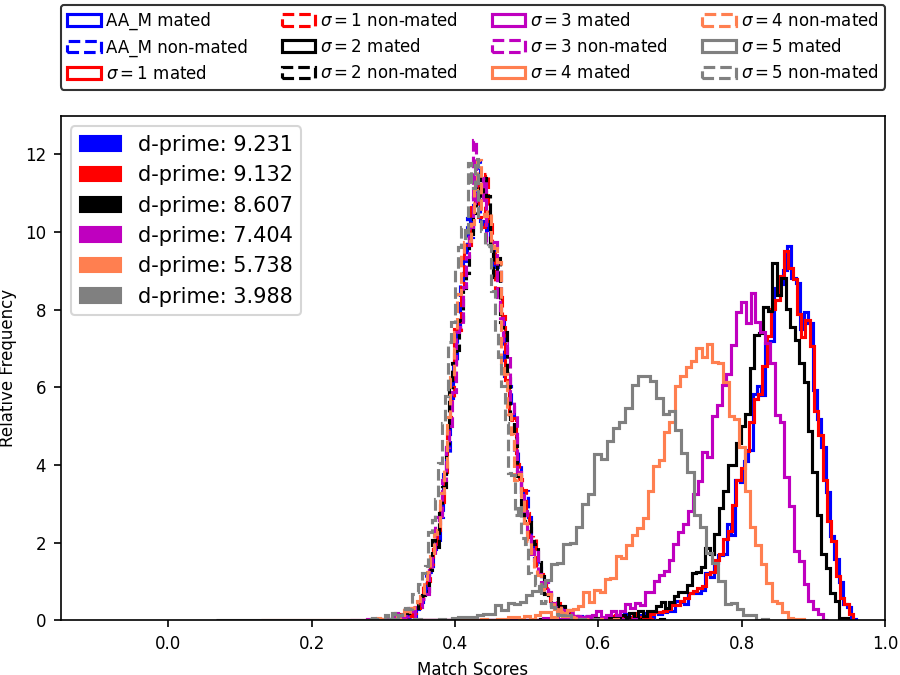}
          \caption{African-American Male}
      \end{subfigure}
      \hfill
      \begin{subfigure}[b]{0.245\linewidth}
        \centering
          \includegraphics[width=1\linewidth]{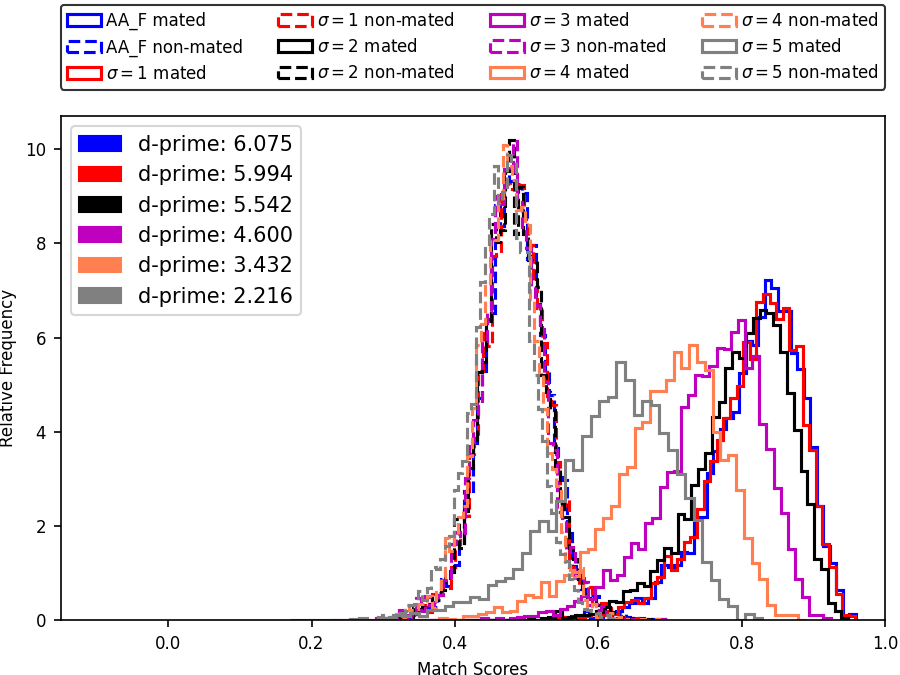}
          \caption{African-American Female}
      \end{subfigure}
  \end{subfigure}
 \caption{Impact of probe image blur on one-to-many matching accuracy. The top row shows the results for ArcFace, the middle row for AdaFace, and the bottom row for MagFace. The impact of blurred probe images remains consistent across the matchers. In all plots, the X-axis represents ``Relative Frequency," while the Y-axis represents ``Match Scores."}
  \vspace{-0.5em}
  \label{fig:gaussian_blur}
\end{figure*}

%% file: fig_texs/downsample_demo.tex
\begin{figure*}[!ht]
  \begin{subfigure}[b]{1\linewidth}
    \centering
      \begin{subfigure}[b]{0.16\linewidth}
        \centering
          \includegraphics[width=1\linewidth]{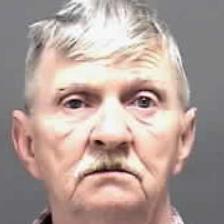}
          \caption{$224 \times 224$}
      \end{subfigure}
      \hfill
      \begin{subfigure}[b]{0.16\linewidth}
        \centering
          \includegraphics[width=1\linewidth]{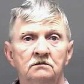}
          \caption{$84 \times 84$}
      \end{subfigure}
      \hfill
      \begin{subfigure}[b]{0.16\linewidth}
        \centering
          \includegraphics[width=1\linewidth]{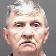}
          \caption{$56 \times 56$}
      \end{subfigure}
      \hfill
      \begin{subfigure}[b]{0.16\linewidth}
        \centering
          \includegraphics[width=1\linewidth]{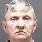}
          \caption{$42 \times 42$}
      \end{subfigure}
      \hfill
      \begin{subfigure}[b]{0.16\linewidth}
        \centering
          \includegraphics[width=1\linewidth]{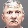}
          \caption{$28 \times 28$}
      \end{subfigure}
  \end{subfigure}
  \caption{Sample probe images with increasing downscaling levels.}
  \vspace{-0.25em}
  \label{fig:downsample_demo}
\end{figure*}

%% file: fig_texs/downsample.tex
\begin{figure*}[t]
\centering
  \begin{subfigure}[b]{1\linewidth}
    \centering
      \begin{subfigure}[b]{0.245\linewidth}
        \centering
          \includegraphics[width=1\linewidth]{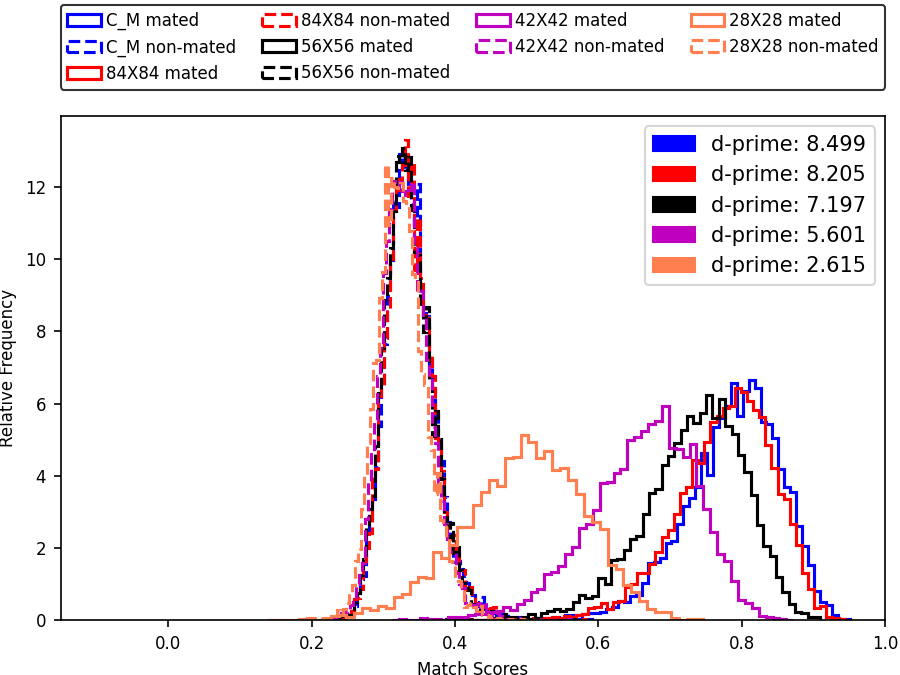}
      \end{subfigure}
      \hfill
      \begin{subfigure}[b]{0.245\linewidth}
        \centering
          \includegraphics[width=1\linewidth]{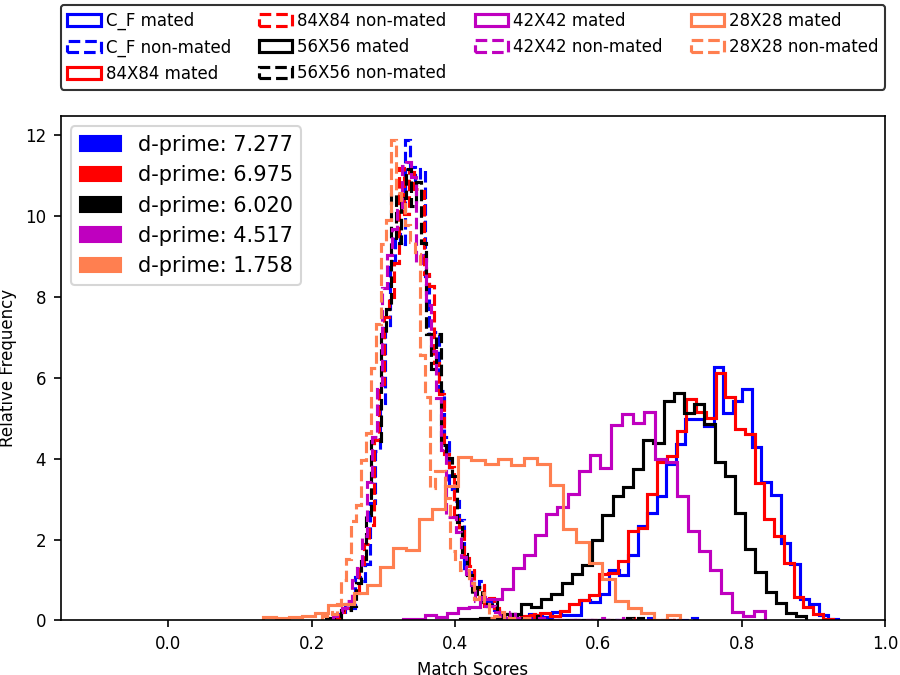}
      \end{subfigure}
      \hfill
      \begin{subfigure}[b]{0.245\linewidth}
        \centering
          \includegraphics[width=1\linewidth]{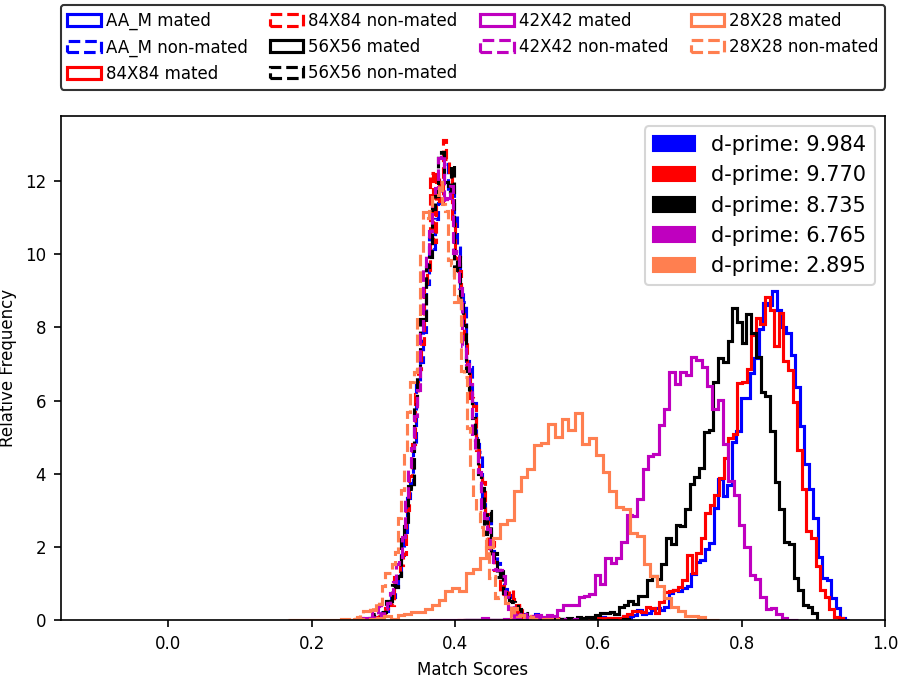}
      \end{subfigure}
      \hfill
      \begin{subfigure}[b]{0.245\linewidth}
        \centering
          \includegraphics[width=1\linewidth]{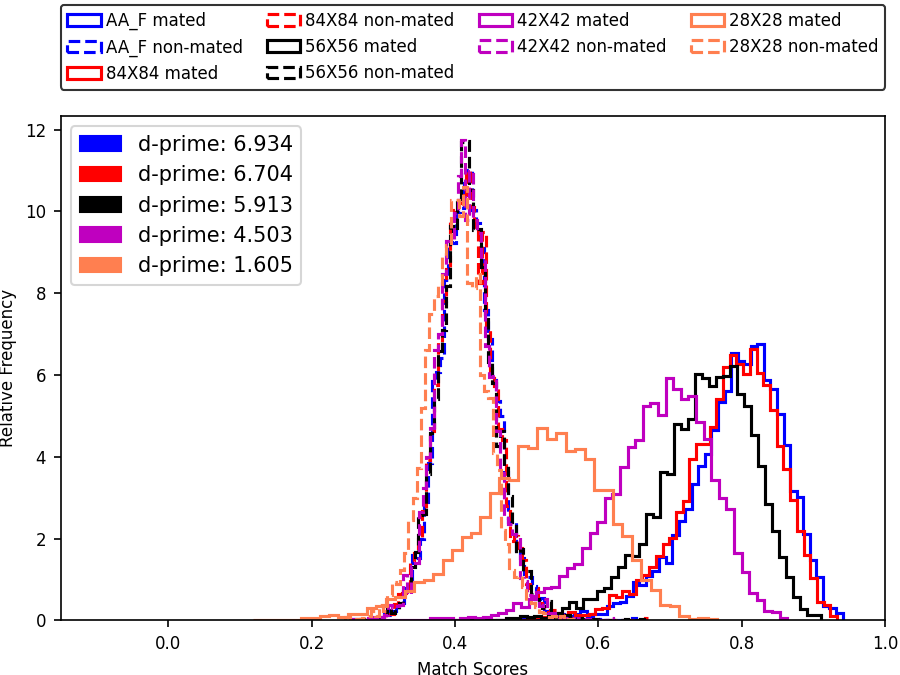}
      \end{subfigure}
  \end{subfigure}

  \vspace{10pt}
  
  \begin{subfigure}[b]{1\linewidth}
    \centering
      \begin{subfigure}[b]{0.245\linewidth}
        \centering
          \includegraphics[width=1\linewidth]{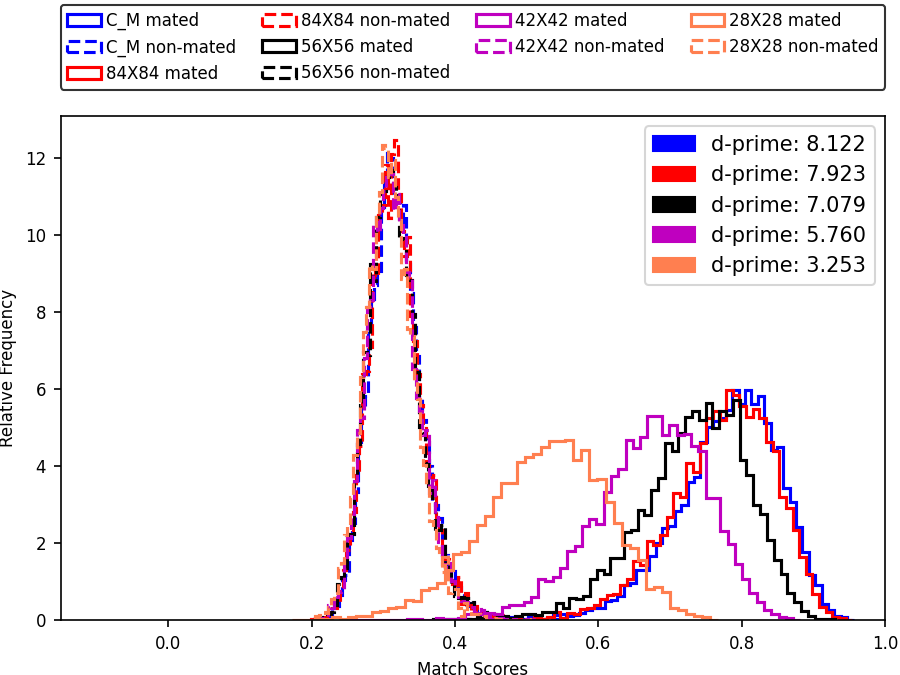}
      \end{subfigure}
      \hfill
      \begin{subfigure}[b]{0.245\linewidth}
        \centering
          \includegraphics[width=1\linewidth]{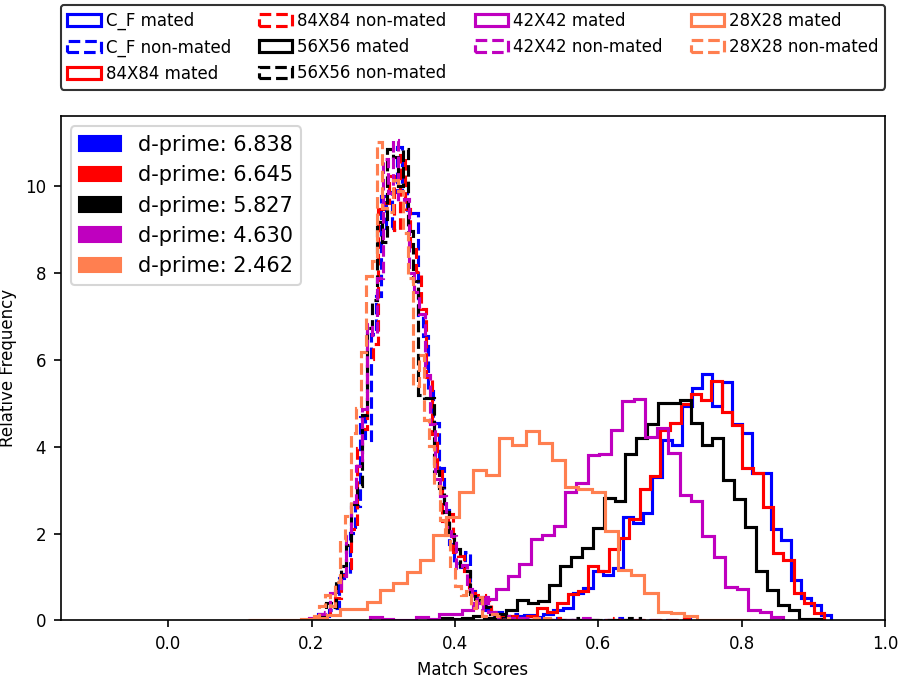}
      \end{subfigure}
      \hfill
      \begin{subfigure}[b]{0.245\linewidth}
        \centering
          \includegraphics[width=1\linewidth]{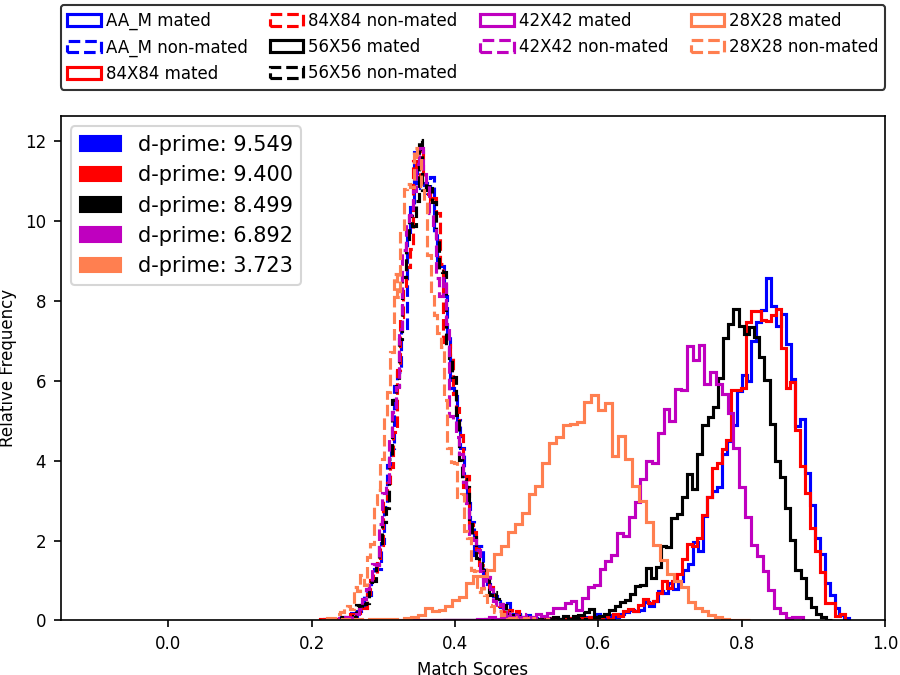}
      \end{subfigure}
      \hfill
      \begin{subfigure}[b]{0.245\linewidth}
        \centering
          \includegraphics[width=1\linewidth]{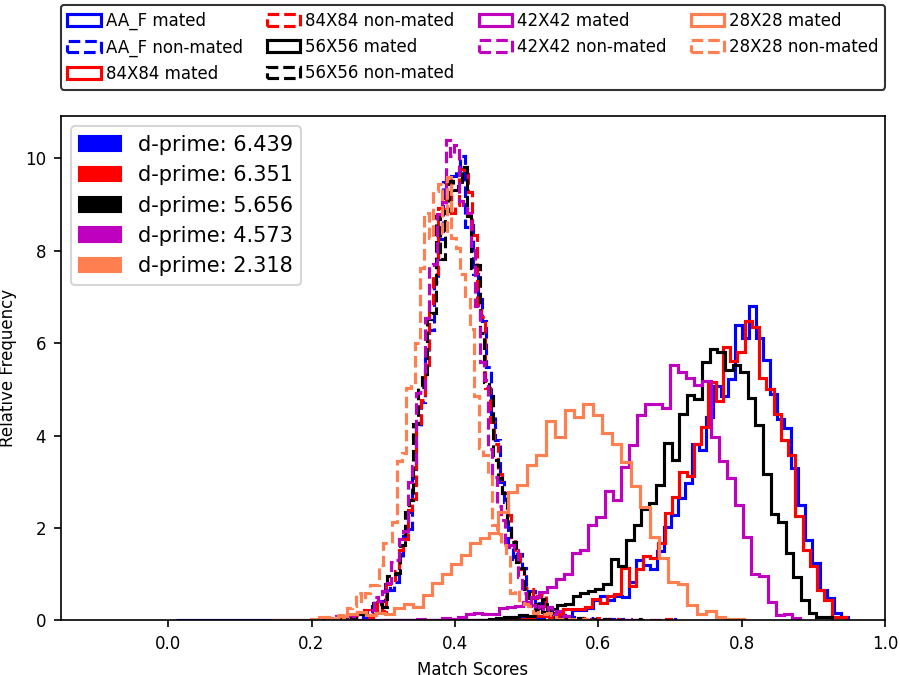}
      \end{subfigure}
  \end{subfigure}

  \vspace{10pt}
  
  \begin{subfigure}[b]{1\linewidth}
    \centering
      \begin{subfigure}[b]{0.245\linewidth}
        \centering
          \includegraphics[width=1\linewidth]{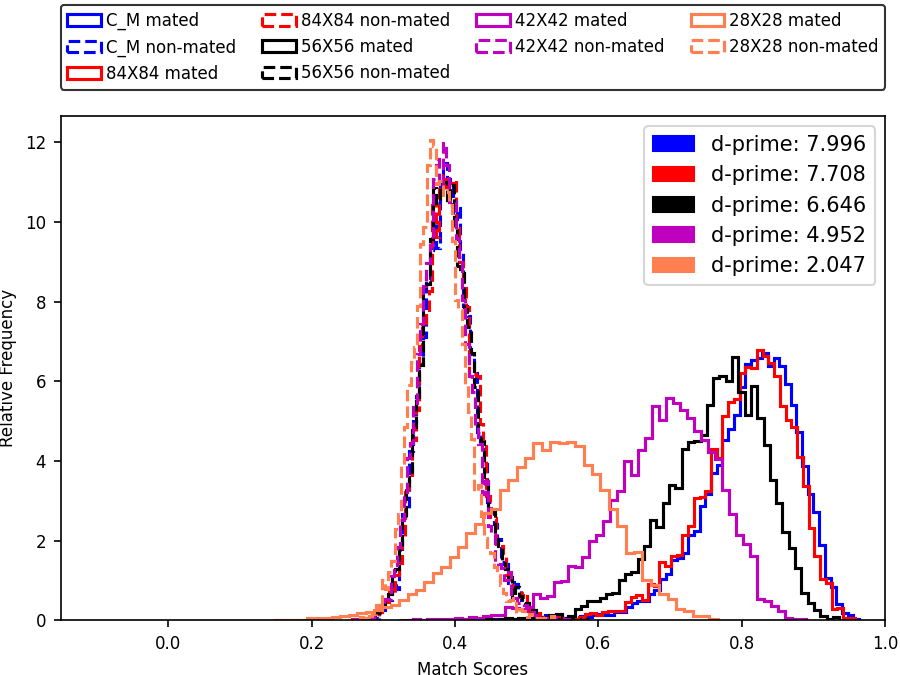}
          \caption{Caucasian Male}
      \end{subfigure}
      \hfill
      \begin{subfigure}[b]{0.245\linewidth}
        \centering
          \includegraphics[width=1\linewidth]{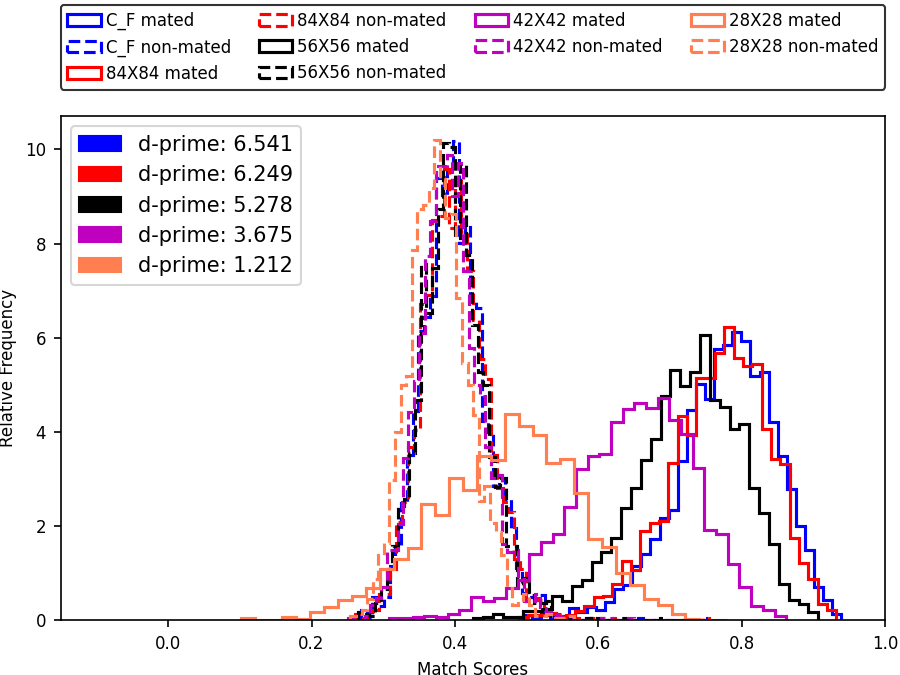}
          \caption{Caucasian Female}
      \end{subfigure}
      \hfill
      \begin{subfigure}[b]{0.245\linewidth}
        \centering
          \includegraphics[width=1\linewidth]{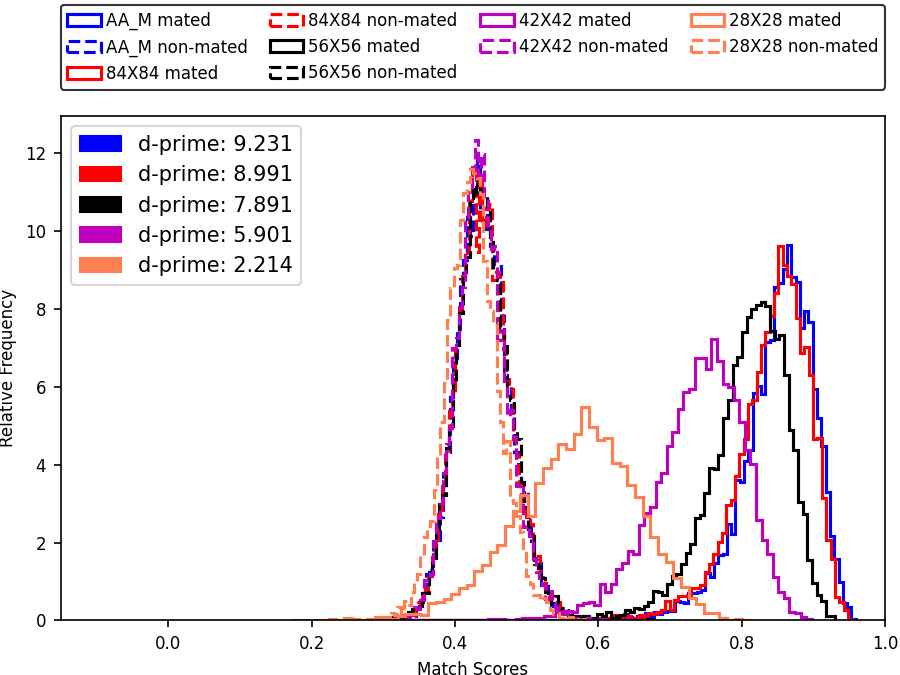}
          \caption{African-American Male}
      \end{subfigure}
      \hfill
      \begin{subfigure}[b]{0.245\linewidth}
        \centering
          \includegraphics[width=1\linewidth]{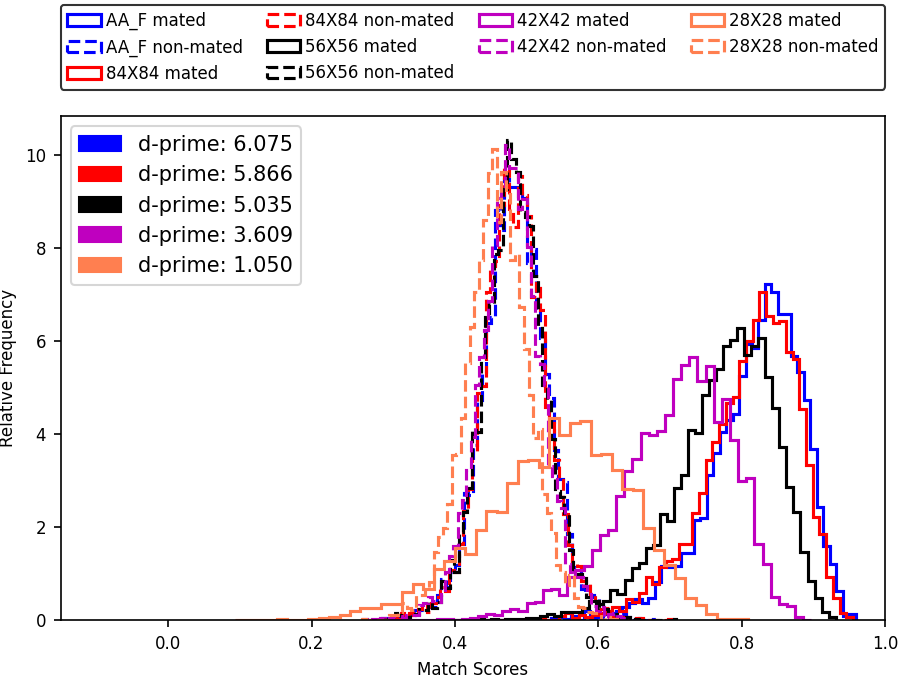}
          \caption{African-American Female}
      \end{subfigure}
  \end{subfigure}
 \caption{Impact of lower-resolution probe on one-to-many matching accuracy. The top row shows the results for ArcFace, the middle row for AdaFace, and the bottom row for MagFace. The impact of downsampled probe images remains consistent across the matchers. In all plots, the X-axis represents ``Relative Frequency," while the Y-axis represents ``Match Scores."}
  \vspace{-1em}
  \label{fig:downsample}
\end{figure*}

%% file: fig_texs/fpir_plots.tex
\begin{figure}[ht!]
\centering
  \begin{subfigure}[b]{1\columnwidth}
    \centering
      \begin{subfigure}[b]{0.495\columnwidth}
        \centering
          \includegraphics[width=1\linewidth]{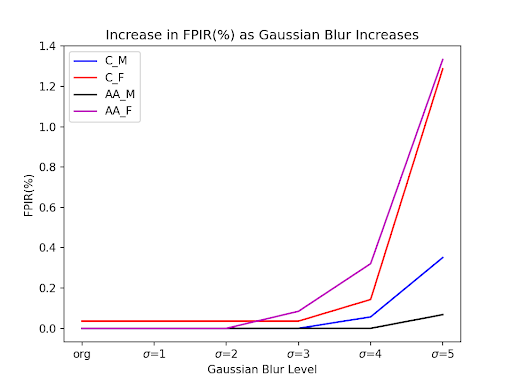}
          \caption{Gaussian Blur}\label{gb}
      \end{subfigure}
      \begin{subfigure}[b]{0.495\columnwidth}
        \centering
          \includegraphics[width=1\linewidth]{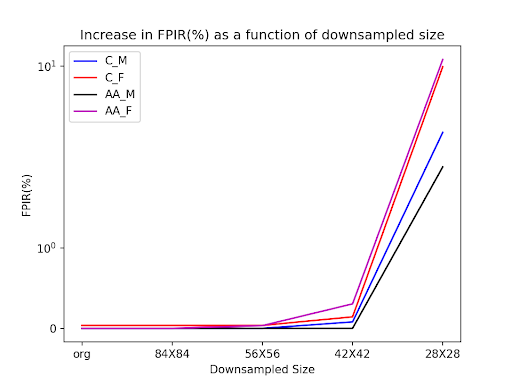}
          \caption{Low Resolution}\label{lr}
      \end{subfigure}
  \end{subfigure}
  \caption{FPIR(\%) for blurred (left) and lower resolution (right) probe images. The plots shown here are for ArcFace. In both plots, the Y-axis represents FPIR(\%), while the X-axis represents the gaussian blur level for the figure on the left and image resolution for the figure on the right. }
  \vspace{-1em}
  \label{fig:gauss_res}
\end{figure}